\documentclass[acmtog,nonacm]{acmart}
\AtBeginDocument{%
  }

\usepackage{multirow}
\usepackage[normalem]{ulem}

\setcopyright{none}
\acmDOI{}
\acmISBN{}

\begin{document}

\title{PDB: Point-Based Deformation Blending for Facial Animation Retargeting}


\author{Sihun Cha}
\email{chacorp@kaist.ac.kr}
\orcid{0000-0001-9506-9438}
\affiliation{%
  \institution{Visual Media Lab, KAIST}
  \city{Daejeon}
  \country{Republic of Korea}
}

\author{Hyeonseung Shin}
\orcid{0009-0004-9539-8978}
\email{hs.shin10@kaist.ac.kr}
\affiliation{%
  \institution{Visual Media Lab, KAIST}
  \city{Daejeon}
  \country{Republic of Korea}
}

\author{Suah Yu}
\orcid{0009-0001-1064-1405}
\email{yusuah@kaist.ac.kr}
\affiliation{%
  \institution{Visual Media Lab, KAIST}
  \city{Daejeon}
  \country{Republic of Korea}
}

\author{Junyong Noh}
\orcid{0000-0003-1925-3326}
\email{junyongnoh@kaist.ac.kr}
\affiliation{%
  \institution{Visual Media Lab, KAIST}
  \city{Daejeon}
  \country{Republic of Korea}
}

\renewcommand{\shortauthors}{Cha et al.}

\newcommand{\bX}{\mathbf{X}}
\newcommand{\bd}{\mathbf{d}}
\newcommand{\br}{\mathbf{r}}
\newcommand{\Ein}{E_{\mathrm{in}}}
\newcommand{\Eout}{E_{\mathrm{out}}}
\newcommand{\Esig}{E_{\mathrm{signal}}}
\newcommand{\Eneu}{E_{\mathrm{neutral}}}

\newcommand{\rev}[1]{{\textcolor{black}{#1}}}
\newcommand{\csh}[1]{{\textcolor{black}{#1}}}
\newcommand{\shs}[1]{{\textcolor{black}{#1}}}
\newcommand{\ysa}[1]{{\textcolor{black}{#1}}}

\begin{abstract}
    Mesh-agnostic facial animation retargeting methods enable expression transfer across meshes with different structures. However, preserving the intended expressions while avoiding local surface artifacts remains challenging: dense per-vertex displacement predictions can exhibit surface noise, whereas the global reconstruction solve used by per-triangle Jacobian methods can introduce undesired coupling between facial regions. To address this, we present PDB, a \textbf{P}oint-Based \textbf{D}eformation \textbf{B}lending method for facial animation retargeting. PDB represents source expressions using a compact set of deformed control points $v$ and distributes their motion over the target surface through blending weights predicted from the target neutral mesh. During inference, the deformed control points are predicted for each source expression, while the blending weights are computed once for a given target neutral mesh and reused across frames. The weights are parameterized with ReLU to enforce non-negativity and allow exact zero entries, followed by row-wise normalization to obtain the normalized blending-weight matrix $w$. Both the deformed control points and blending weights are predicted without a predefined control structure or precomputed coordinates. The target mesh is reconstructed directly as $wv$, without a learned per-vertex or per-triangle deformation decoder or a global reconstruction solve. Trained with self-retargeting reconstruction supervision, PDB supports cross-identity retargeting without paired cross-identity training expressions. Experiments demonstrate accurate retargeting and fast inference, with localized influence observed in the learned weights. Joint evaluation of expression accuracy and local surface preservation shows that PDB retains the intended facial motion while reducing surface artifacts relative to the evaluated dense displacement method. Perceptual evaluations further support its expression fidelity and visual quality in both self- and cross-retargeting.

\end{abstract}



\begin{teaserfigure}
  \includegraphics[width=\textwidth]{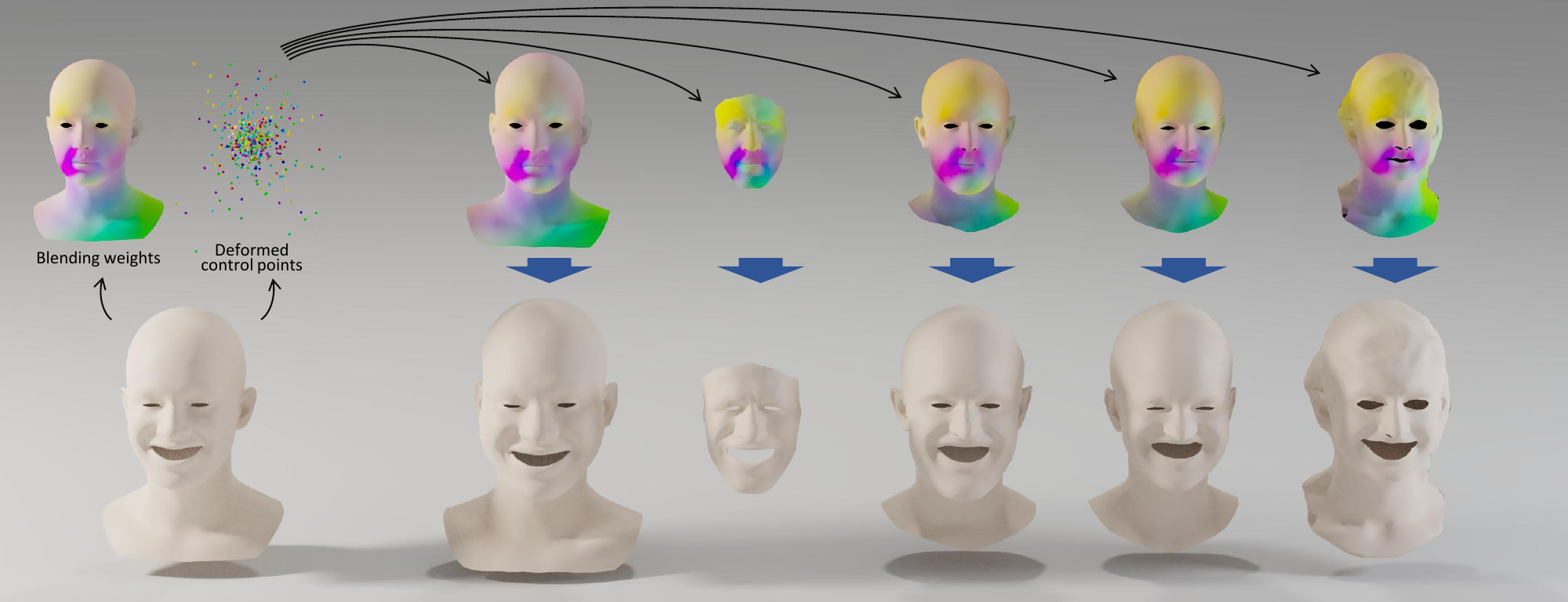}
  \caption{PDB retargets facial expressions by combining deformed control points predicted from the source expression with localized weights predicted from the target shape, enabling direct retargeting across diverse identities.
  The weights are visualized by assigning a random color to each control point and blending according to the predicted weight. (face models from BIWI~\cite{fanelli2013biwi}, ICT~\cite{li2020ictface}, COMA~\cite{ranjan2018coma} and Multiface~\cite{wuu2022multiface})
  }
  \label{fig:teaser}
\end{teaserfigure}

\maketitle

\section{Introduction}
\label{sec:intro}

Deep learning has significantly improved the quality and controllability of 3D facial animation, enabling realistic motion transfer and expression synthesis. Recently, mesh-agnostic approaches have extended these advances to face meshes with different resolutions, triangulations, and geometric structures~\cite{wang2023zpt,chandran2022shape,qin2023NFR,nocentini2024scantalk,cha2025nfs}. A key challenge in this setting is how to represent deformation so that local facial motion is preserved while avoiding surface artifacts and irregular geometry.

\begin{figure}[t]
\begin{center}
\includegraphics[width=\linewidth]{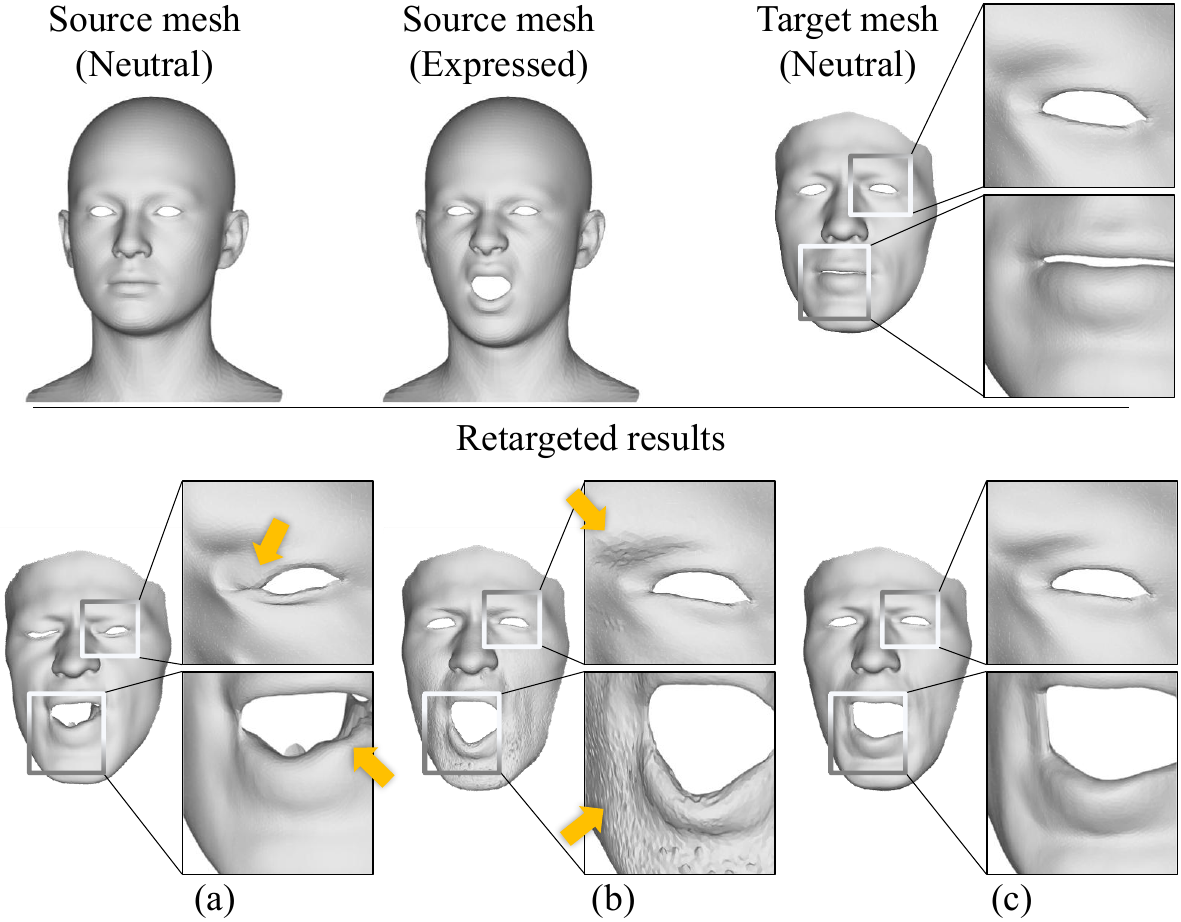}
\end{center}
\caption[teaser]{
    Comparison of our method with previous approaches. Facial retargeting results from (a) per-triangle Jacobian prediction method~\cite{qin2023NFR}, (b) per-vertex displacement prediction method~\cite{cha2025nfs}, and (c) ours using a high-resolution mesh from the BIWI~\cite{fanelli2013biwi} dataset (number of vertices: $23,370$). The yellow arrow indicates the artifacts.
}
\label{fig:teaser_compare}
\end{figure}

Existing methods mostly represent deformation on dense surface elements, such as per-vertex displacements or per-triangle Jacobians. Per-vertex approaches~\cite{wang2023zpt,chandran2022shape,cha2025nfs} predict local motion and efficiently capture fine-scale expressions, but \csh{can produce local surface artifacts} on dense meshes, as shown in Figure~\ref{fig:teaser_compare} (b). Jacobian-based methods~\cite{aigerman2022njf,qin2023NFR} instead recover vertex positions through a global solve, producing spatially smooth and regular results. 
\csh{However, the global solve can introduce undesired coupling across distant regions, as illustrated in Figure~\ref{fig:teaser_compare} (a).}
\csh{These observations motivate a deformation representation that preserves local facial motion and surface structure without requiring a reconstruction process via global solve.}

To address this, we propose PDB, a point-based deformation \csh{blending} method for mesh agnostic facial animation retargeting.
\csh{PDB predicts a compact set of deformed control points \(v\) from the source neutral--expression pair and blending weights \(w\) from the target neutral mesh. The target expression is reconstructed by their direct weighted combination \(wv\), where each row of \(w\) defines the contribution of the control points to a vertex of the target mesh.
}

\csh{PDB learns to predict the deformed control points and the blending weights from self-retargeting reconstruction supervision, without a predefined control structure or precomputed coordinates. For cross-identity retargeting, PDB combines the deformed control points predicted from the source with blending weights predicted from another identity's neutral mesh. Across the evaluated identities, corresponding weight columns exhibit concentrated influence on similar facial regions, providing empirical evidence of a shared spatial organization for expression transfer, as illustrated in Figure~\ref{fig:teaser}.}


\csh{ReLU enforces non-negativity and allows exact zero weights, while row-wise normalization controls the total blending weight at each target vertex. Spatial locality is assessed through the learned weight distributions and the resulting deformations. During inference, the deformed control points are predicted for each source expression, while the blending weights are computed once for a given target neutral mesh and reused across frames. Unlike previous approaches~\cite{qin2023NFR,cha2025nfs} that evaluate a learned deformation decoder at every target vertex or triangle, PDB reconstructs the target mesh through a single matrix multiplication $wv$. We refer to this output formulation as decoder-free because it requires neither per-element neural decoding nor a global reconstruction solve.}


\csh{Experiments evaluate PDB on self-retargeting and cyclic retargeting, together with neutral reconstruction, learned weight support, and qualitative cross-identity transfer. Joint analysis of expression accuracy and local surface preservation shows that PDB preserves the intended facial motion while reducing surface artifacts relative to the evaluated dense displacement method. Perceptual evaluations further support its expression fidelity and visual quality. Runtime comparisons demonstrate shorter end-to-end sequence processing times than the compared methods, including target-specific preprocessing.}

\section{Related work}
\label{sec:related_work}

\subsection{Facial Animation Retargeting}

Facial animation retargeting has long been studied as a problem of transferring expression deformation from one face to another since the foundational work by~\citet{parke1972computer}. Early methods relied on dense correspondences between source and target meshes~\cite{noh2001expression,sumner2004deformation}, while later approaches reduced this requirement by using sparse correspondences together with interpolation schemes~\cite{bickel2007multi,ma2008facial}. With the widespread use of blendshape models, many retargeting methods were developed in this space, including radial-basis-function-based interpolation~\cite{bickel2007multi,orvalho2008transferring,ribera2017facial,umenhoffer2012facial}, velocity-domain transfer~\cite{seol2012spacetime}, and segment-based localized transfer~\cite{liu2011framework}. For a broader overview of motion retargeting techniques, we refer readers to the recent survey by \citet{zhu2024review}. 

Deep learning reformulated the problem by learning deformation representations directly from data. Early neural approaches learned latent spaces of facial expression using graph-based or autoencoding architectures~\cite{ranjan2018coma,bouritsas2019neural3dmm,groueix20183d,tan2018variational}. Later works improved retargeting quality by disentangling identity and expression in the latent space~\cite{chandran2020semantic,jiang2019disentangled}, while others addressed retargeting from weakly paired or unpaired data~\cite{gao2018automatic}. These methods demonstrated that transferable deformation priors can be learned without explicit hand-crafted correspondences. Instead, deformation is often expressed through latent codes that are not directly interpretable.

Recent mesh-agnostic methods generalized retargeting to arbitrary mesh structures by predicting local deformation quantities defined on surface elements, such as per-vertex displacements~\cite{chandran2022shape,wang2023zpt,cha2025nfs} or per-triangle Jacobians~\cite{aigerman2022njf,qin2023NFR}. Per-vertex approaches can capture subtle expressions and fine-scale motion, but they are sensitive to prediction noise on dense meshes. Jacobian-based methods reconstruct vertex positions through a global solve, which often improves surface regularity but introduces global coupling and additional computational overhead. In contrast, our method does not predict deformation directly on mesh vertices or triangles. Instead, it represents facial deformation through deformed control points and localized weights, and reconstructs the target mesh directly from their matrix product. NFS~\cite{cha2025nfs} also uses spatially localized skinning weights, but these weights condition a neural decoder that predicts per-vertex deformation. In PDB, the weights are themselves one of the two explicit geometric factors and directly reconstruct the target mesh through $wv$, without a deformation decoder.

\subsection{Control-Based Deformation}

\csh{Control-based deformation propagates motion from a compact set of controls to dense geometry through spatially varying weights. Linear blend skinning (LBS)~\cite{magnenat1988LBS} blends transformations associated with skeletal controls using per-vertex skinning weights. Compressed Skinning for Facial Blendshapes (CSFB)~\cite{kavan2024compskin} converts facial blendshapes into an efficient linear blend skinning representation. PDB shares the motivation of compact animation representation, but learns to predict source-dependent deformed control points and target-dependent blending weights for cross-identity retargeting. Cage-based deformation (CBD) propagates the motion of cage vertices to an embedded shape through geometric coordinate functions, with representative formulations including mean value coordinates~\cite{ju2005mvctri} and harmonic coordinates~\cite{joshi2007hc}. For a comprehensive overview of CBD, we refer readers to the survey by \citet{stroter2024surveyCBD}.}

Several recent works combine control-based deformation with deep learning for editing and deforming arbitrary meshes~\cite{yifan2020neuralcage,jakab2021keypointdeformer,liu2021deepmetahandles,dodik2023vbc}. NeuralCage~\cite{yifan2020neuralcage} predicts cage offsets to deform a source mesh toward a target shape. KeypointDeformer~\cite{jakab2021keypointdeformer} learns sparse keypoints that serve as deformation handles. \csh{DeepMetaHandles~\cite{liu2021deepmetahandles} learns disentangled meta-handles as combinations of prescribed control-point translations and uses precomputed biharmonic coordinates to propagate them over the mesh. VBC~\cite{dodik2023vbc} parameterizes continuous, valid barycentric-coordinate functions using a neural field, allowing the coordinates to be optimized for smoothness and deformation-aware energies.} NeuralMLS~\cite{shechter2022neuralmls} predicts weights from sparse control points and uses an MLS solver to propagate their motion to a dense shape.
\csh{PDB predicts both deformed control points and blending weights without a prescribed cage or precomputed geometric coordinates.}

\csh{Localized deformation representations have also been explored for facial animation and editing. SPLOCS~\cite{neumann2013splocs} decomposes registered mesh animation sequences into sparse, spatially localized per-vertex components for intuitive animation editing, including facial animations. Whereas SPLOCS represents animation through fixed per-vertex deformation components with varying activation coefficients, PDB predicts expression-dependent 3D deformed control points and distributes their motion through blending weights predicted from the target neutral mesh. CUBE~\cite{chandran2026cube} represents 3D faces using a lattice of high-dimensional control features that are locally blended by B-spline bases and decoded through a residual MLP over a fixed template domain. Its local support enables localized facial editing, while differences between control features can be transferred across identities to reproduce expressions. PDB blends explicit 3D deformed control points directly into target vertex positions, without an additional learned decoder after blending. Unlike the local support provided by CUBE's B-spline bases, spatial locality in PDB is learned and evaluated empirically.}

\section{Method}
\begin{figure*}[t]
\includegraphics[width=\linewidth]{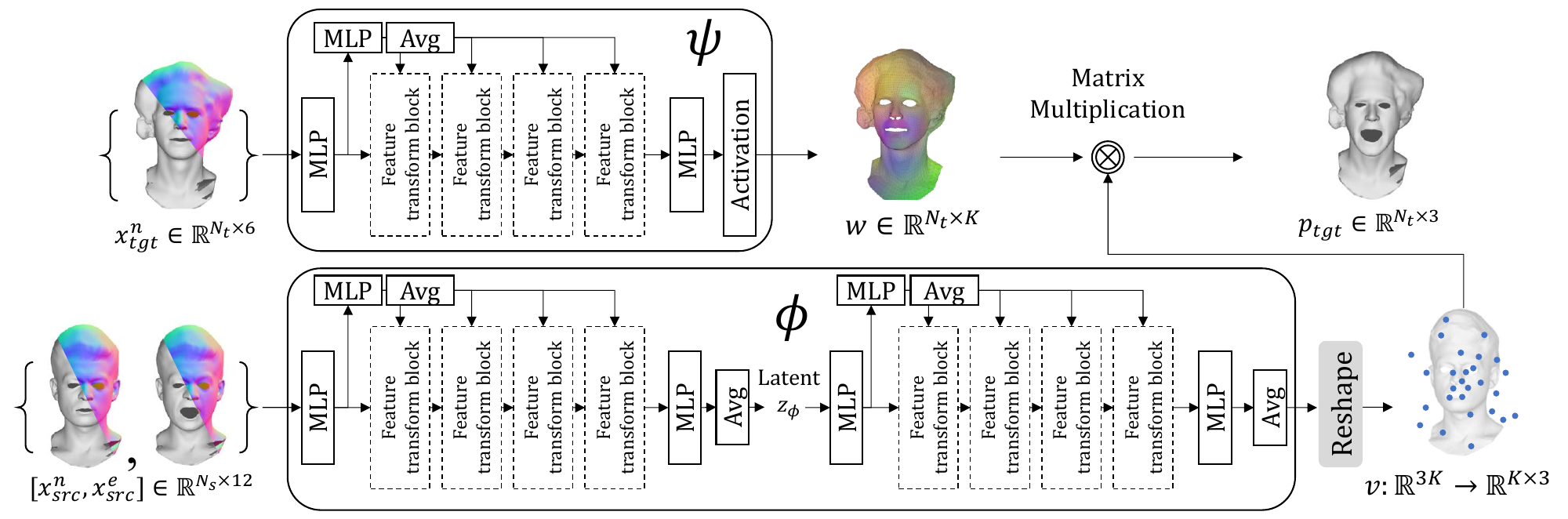}
\caption[Method overview]{
    Method overview for cross-identity retargeting at inference.
}
\label{fig:architecture_detail}
\end{figure*}

In this section, we describe our method following the forward data flow of the system. Figure~\ref{fig:architecture_detail} shows the overall pipeline. Given a source neutral mesh, a source expression mesh, and a target neutral mesh, the goal is to produce the target expression mesh. Instead of predicting deformation directly on target vertices or triangles, our method represents facial deformation using two explicit geometric factors: deformed control points and localized weights. The deformed control points are predicted from the source expression, and the localized weights are predicted from the target identity. The target expression is then reconstructed directly by their weighted combination. For a fixed target, the weights are computed once and reused across source frames.

\subsection{Overview and Reconstruction}

 PDB consists of two encoders: an expression point encoder $\phi$ and an identity weight encoder $\psi$. From the input mesh, we extract a per-vertex feature \(x\in\mathbb{R}^{6}\) consisting of the vertex position and the vertex normal. 
The expression point encoder \(\phi\) takes the row-wise concatenated source features \([x_{\mathrm{src}}^{n},x_{\mathrm{src}}^{e}] \in \mathbb{R}^{N_s\times 12}\) as input, where \(x_{\mathrm{src}}^{n}\) and \(x_{\mathrm{src}}^{e}\) are the features of the source neutral and source expression meshes, respectively. The source neutral and expression meshes must have corresponding vertices for this concatenation; no vertex correspondence between the source and target meshes is required. The encoder predicts a set of deformed control points:
\begin{equation}
\label{eq:cage_vertex_prediction}
v=\phi([x_{\mathrm{src}}^{n},x_{\mathrm{src}}^{e}]),
\end{equation}
where \(v\in\mathbb{R}^{K\times 3}\) contains \(K\) deformed control points \(v_i\in\mathbb{R}^{3}\). \(K\) can be adjusted as a hyperparameter before training, and the network always predicts \(K\) control points after training.

The identity weight encoder \(\psi\) takes the target neutral features \(x_{\mathrm{tgt}}^{n}\in\mathbb{R}^{N_t\times 6}\) as input and predicts a weight matrix
\begin{equation}
\label{eq:coordinate_prediction}
w=\psi(x_{\mathrm{tgt}}^{n}),
\end{equation}
where \(w\in\mathbb{R}^{N_t\times K}\) contains the final non-negative blending weights, including the normalization steps described in Section~\ref{sec:weight_formulation}. Each element \(w_i^j\) denotes the contribution of the \(i\)-th control point to the \(j\)-th target vertex. The target expression is reconstructed as
\begin{equation}
\label{eq:method_init}
\hat{p}_{\mathrm{tgt}}^{j}=\sum_{i=1}^{K} w_i^j v_i.
\end{equation}
Equivalently, for all target vertices,
\begin{equation}
\label{eq:matrix_form}
\hat{P}_{\mathrm{tgt}} = wv,
\end{equation}
where \(\hat{P}_{\mathrm{tgt}}\in\mathbb{R}^{N_t\times 3}\) denotes the predicted target expression mesh.

\begin{figure}[t]
\begin{center}
\includegraphics[width=\linewidth]{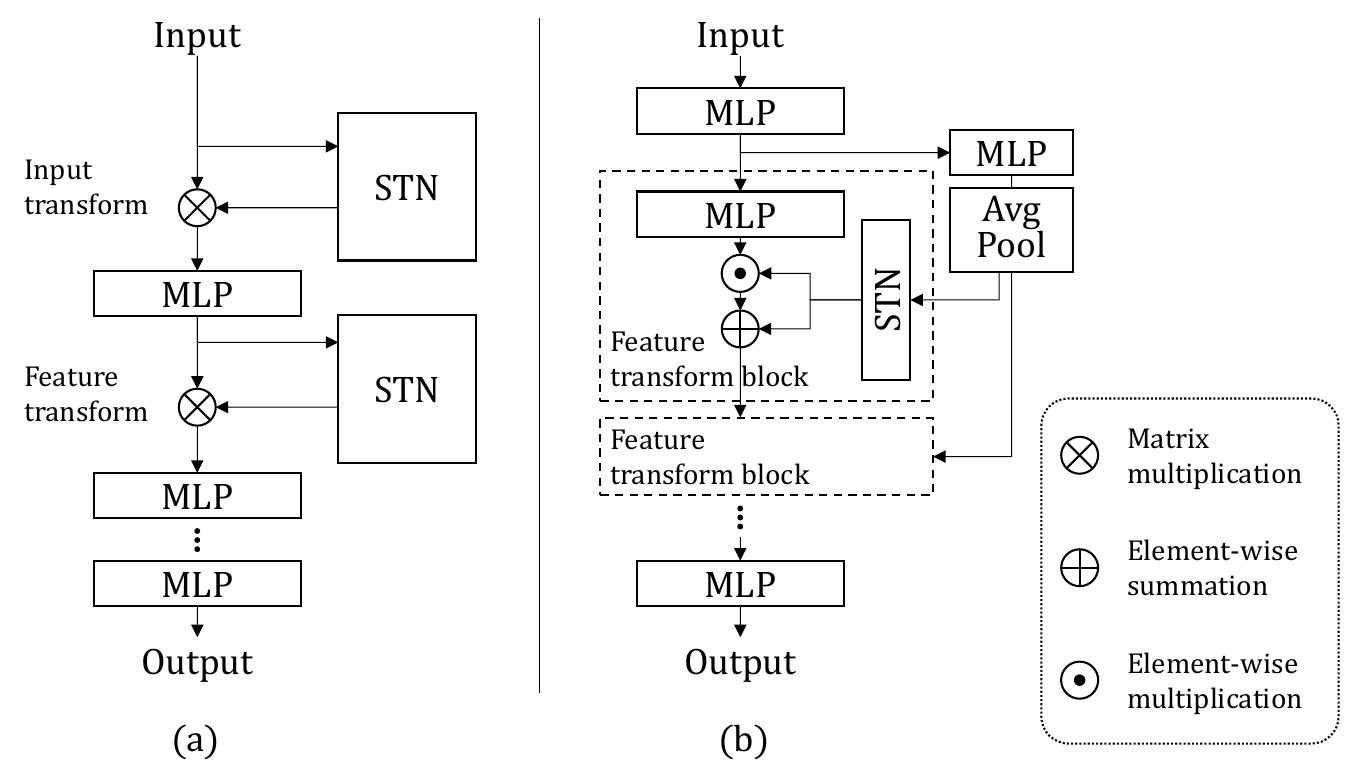}
\end{center}
\caption[Illustration of the feature transform block]{
    Illustration of the feature transform block (b) compared with the spatial transform in PointNet~\cite{qi2017pointnet} (a). 
}
\label{fig:architecture}
\end{figure}

\subsection{Blending-Weight Parameterization}
\label{sec:weight_formulation}

\csh{The blending weights determine how the predicted deformed control points contribute to each target vertex. We parameterize the weights using column-wise normalization and ReLU, followed by row-wise normalization to obtain the final blending weights used for reconstruction.}

\paragraph{Non-negative parameterization.}
Let $z\in\mathbb{R}^{N_t\times K}$ denote the output of the final linear layer of $\psi$. We normalize each column over the target vertices and apply ReLU:
\begin{equation}
\tilde{w}_i=\operatorname{ReLU}\left(
\frac{z_i}{\max(\|z_i\|_2,\varepsilon)}
\right),
\end{equation}
where $z_i,\tilde{w}_i\in\mathbb{R}^{N_t}$ denote the $i$-th columns and ReLU acts element-wise. We use $\varepsilon=10^{-12}$. The denominator floor keeps zero-norm columns well-defined. ReLU enforces non-negativity and allows exact zero entries.

\paragraph{Row-wise normalization.}
For each target vertex, the weights are normalized row-wise as follows:
\begin{equation}
\label{eq:pou}
w_i^j
=
\csh{\frac{\tilde{w}_i^j}{\sum_k \tilde{w}_k^j+\varepsilon},}
\end{equation}
\csh{where $\varepsilon=10^{-12}$ is used for numerical stability. For rows whose total weight is sufficiently larger than $\varepsilon$, the normalized weights sum approximately to one. An all-zero row remains zero and reconstructs its vertex at the origin; this parameterization does not impose a separate nonzero-support constraint.}

\csh{The control-point predictions and blending weights are learned jointly through the reconstruction objective defined in Equation~\ref{eq:loss_recon_def}.}

\paragraph{Sparse support and spatial locality.}
\csh{Facial expressions often affect localized regions, as reflected by the sparse spatial support of many delta blendshapes~\cite{lewis2014practice}. Exact zero weights exclude individual control-point contributions at a target vertex, but do not by themselves enforce spatial locality. Although spatial locality is not explicitly constrained, the learned weights exhibit localized support, as shown in Figure~\ref{fig:compare_activation}.}

\subsection{Architecture}\label{sec:network_architecture}
\csh{
The networks are based on a Multi-Layer Perceptron (MLP). Inspired by PointNet~\cite{qi2017pointnet}, which applies feature-space transformations via spatial transformer networks (STN), we design a feature transform block. The original PointNet transform models a linear map without bias:
\begin{equation}
    y = xT,
\end{equation}
where $x, y \in \mathbb{R}^{N \times F}$ are the features before and after transformation, $T \in \mathbb{R}^{F \times F}$ is the predicted transform, $N$ is the number of vertices, and $F$ is the feature dimension. In contrast, our block predicts a feature-wise affine transformation with learned scales and biases. We implement it as an element-wise affine map:
\begin{equation}
    y = a \odot x + b,
\end{equation}
where $a$ and $b \in \mathbb{R}^{1{\times}F}$ are the predicted per-element weights and biases, respectively, and $\odot$ denotes Hadamard (element-wise) multiplication. Note that $a$ and $b$ are broadcast to match the dimension size $N{\times}F$ before the multiplication.
Figure~\ref{fig:architecture} illustrates the feature transform block.}

\csh{
The \emph{identity weight encoder} $\psi$ uses an input linear projection, four feature transform blocks, and an output linear projection. An auxiliary MLP branch produces a pooled transform code that conditions the feature-wise scales and biases. The hidden MLPs use layer normalization and \textit{LeakyReLU} activations. The output projection is followed by the weight parameterization in Section~\ref{sec:weight_formulation}.
The \emph{expression point encoder} $\phi$ follows the same feature-transform pattern in two successive encoder stages, with an average-pooling layer and no activation at the output. Figure~\ref{fig:architecture_detail} shows the full architecture.
}

\subsection{Training}

Following~\citet{qin2023NFR}, all meshes are standardized to a common orientation and scale before training.
\csh{
PDB is trained only with self-retargeting supervision; no paired cross-identity ground-truth expressions are used. During training, the source neutral mesh, source expression mesh, and target neutral mesh belong to the same identity, and the corresponding expression mesh provides the reconstruction target. Cross-identity retargeting is obtained at inference time by combining the source-dependent control points predicted by $\phi$ with the target-dependent weights predicted by $\psi$.
}

The reconstruction objective supervises the expression in the inner face and preserves the neutral geometry in the surrounding region:
\begin{equation}
\label{eq:loss_recon_def}
\begin{split}
\mathcal{L}_{\mathrm{recon}}
&=\frac{1}{N_t}\left\|M_{\mathrm{tgt}}^{\mathrm{in}}\odot
(\hat{P}_{\mathrm{tgt}}^{e}-P_{\mathrm{GT}}^{e})\right\|_F^2\\
&\quad+\frac{1}{N_t}\left\|M_{\mathrm{tgt}}^{\mathrm{out}}\odot
(\hat{P}_{\mathrm{tgt}}^{e}-P_{\mathrm{GT}}^{n})\right\|_F^2.
\end{split}
\end{equation}
where $\|\cdot\|_F$ is the Frobenius norm and $\odot$ denotes element-wise multiplication with each vertex mask broadcast over its three coordinates. The factor $1/N_t$ averages the squared Euclidean residuals over all target vertices. Because the masks multiply the residuals before squaring, their effective weights in the squared loss are $(M^{\mathrm{in}})^2$ and $(M^{\mathrm{out}})^2$. Here, $\hat{P}_{\mathrm{tgt}}^{e}$ denotes the predicted target expression mesh, and $P_{\mathrm{GT}}^{e}$ and $P_{\mathrm{GT}}^{n}$ respectively denote the ground-truth target expression and neutral mesh. $M_{\mathrm{tgt}}^{\mathrm{in}}$ and $M_{\mathrm{tgt}}^{\mathrm{out}}$ denote the masks for the inner facial region and its complement, respectively.

\begin{figure}[t]
\begin{center}
\includegraphics[width=\linewidth]{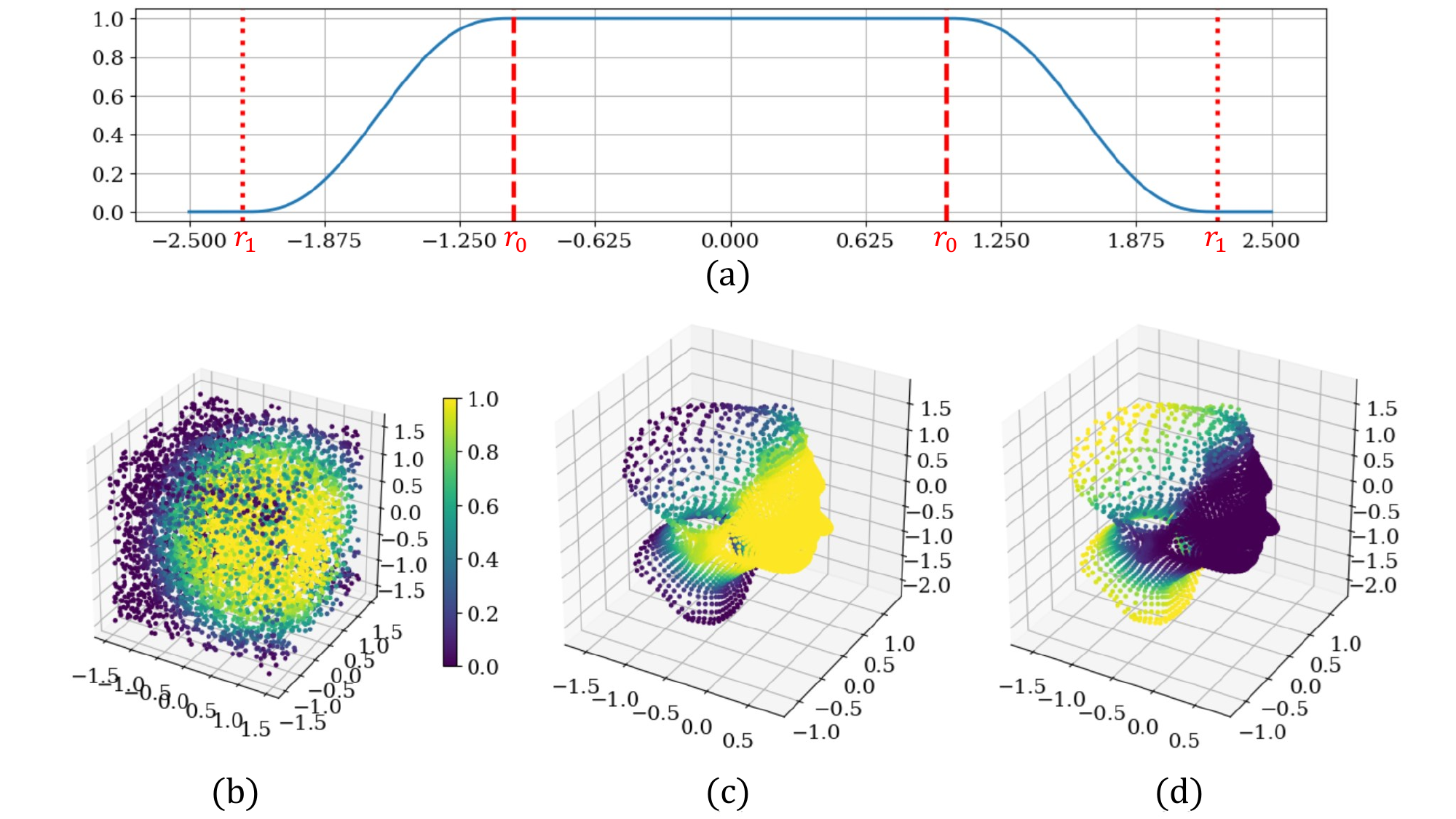}
\end{center}
\caption[Visualization of the hat function]{
    Visualization of the hat function in 1D points (a), values on random sampled 3D points (b), and the obtained masks $M^{\text{in}}$ (c), and $M^{\text{out}}$ (d).
    $r_0$ and $r_1$ are indicated with red dashed lines and dotted lines in (a), respectively.
}
\label{fig:hat_function}
\end{figure}

The masks are used to emphasize expression learning on the inner face while preserving the surrounding region. These masks are constructed using a hat function $h:\mathbb{R}^3\to[0,1]$ centered at the nose peak vertex as follows:
\begin{align}
\label{eq:hat_function}
h(p)&=
\begin{cases}
1, & r(p)\le r_0,\\[2pt]
1 - S\left(\dfrac{r(p)-r_0}{r_1-r_0}\right), & r_0<r(p)<r_1,\\[4pt]
0, & r(p)\ge r_1,
\end{cases}\\[4pt]
r(p)&=\|p-c\|_2, \\[4pt]
S(t)&=10t^3-15t^4+6t^5,
\end{align}
where $(r_0,r_1)=(1.0,\,2.25)$ and $c$ is the nose peak vertex in the standardized coordinate system.
We then define the inner and outer face masks for the vertices of the target neutral mesh $P_\text{tgt}$ as follows:
\begin{equation}
M^{\text{in}} = [h(p_{tgt}^1),\dots,h(p_{tgt}^{N_t})]^\top, \quad
M^{\text{out}} = \mathbf{1} - M^{\text{in}},
\end{equation}
where each entry specifies a continuous mask value at a target vertex. The masks overlap in the transition region $r_0<r(p)<r_1$; they are not disjoint binary regions.
Figure~\ref{fig:hat_function} visualizes the hat function and the resulting masks.

\section{Experiments}

To assess our design choices, we analyzed the effects of the activation function, POU constraint, hat function mask, and deformation representation (Section~\ref{sec:design_study}). 
To evaluate retargeting accuracy, we compared our method, PDB, with state-of-the-art facial animation retargeting methods on self-retargeting and cyclic-retargeting tasks (Section~\ref{sec:precision_qual}), using Mean Squared Error (MSE) as the evaluation metric.
Self-retargeting transfers a source expression to the neutral mesh of the same identity and is therefore evaluated as reconstruction. Cyclic-retargeting transfers the source expression to a different target identity and then back to the source neutral mesh, allowing error measurement without ground-truth cross-identity targets.
Cycle error measures round-trip stability on unseen identities but does not by itself guarantee the semantic correctness of the intermediate cross-identity expression. We therefore report it together with intermediate qualitative results and the perceptual user study.

We evaluate expression reconstruction with the continuous inner-face mask $M^{\mathrm{in}}$ and neutral-geometry preservation with its complement $M^{\mathrm{out}}$. Each metric averages the squared Euclidean masked residual over target vertices, following the corresponding term in Equation~\ref{eq:loss_recon_def}. We use the continuous masks defined in the Method section with $(r_0,r_1)=(1.0,2.25)$; the two masks overlap in the transition band rather than defining disjoint binary regions.
To evaluate perceptual quality, we conducted a user study to judge the perceptual quality of the results (Section~\ref{sec:user_study}).
To evaluate local surface preservation, we measured the MSE between the Laplacian coordinates of the predicted and ground-truth meshes on the inner facial region (Section~\ref{sec:surface_preservation}). Note that we use `Ours` to denote the design variants of our model, and `PDB` to denote the final model.
Please refer to the supplementary material for additional application results.

For each experiment, we trained the model using the AdamW optimizer (betas=($0.9, \,0.999$), weight decay=$0.01$), with a learning rate of $1\times$$10^{-4}$, and a batch size of $16$ using a single NVIDIA-RTX-A5000 GPU. 
\csh{For design study (Section~\ref{sec:design_study}), open-source benchmark data (COMA~\cite{ranjan2018coma}, BIWI~\cite{fanelli2013biwi}, and Multiface~\cite{wuu2022multiface}) were used. Each model was trained for $200$ epochs. We split the identities into train/validation/test sets of 9/1/2 for COMA, 6/6/8 for BIWI, and 10/1/1 for Multiface, where the training and validation identities were the same for BIWI.
For the main comparison (Section~\ref{sec:precision_qual}), we use ICT-FaceKit and Multiface. Following the protocol of NFS, the 111 ICT identities are divided into 100/1/10 training, validation, and test identities, while the 13 Multiface identities are divided into 11/1/1 identities. Each model was trained until the loss converged. ICT and Multiface samples are uniformly sampled per batch. All compared methods use the same identity splits, selected frames and alignment process.}

\subsection{Design Study}\label{sec:design_study}

In this section, we validate our design choices. For the base setting, we set the number of control points to $K\text{=}512$ in Equation~\ref {eq:method_init}, applied \textit{ReLU} activation for the identity weight encoder $\psi$ with explicit row-wise normalization, and only a reconstruction loss $\mathcal{L}_{\mathrm{recon}}$ is applied. The base setting is indicated as `Ours$^\dagger$'.
We varied each component to assess the respective effect.
The results are shown in Table~\ref{tab:design_study} and Figure~\ref{fig:design_study}.

\subsubsection{Activation variants}

\begin{figure}[h]
\begin{center}
\includegraphics[width=\linewidth]{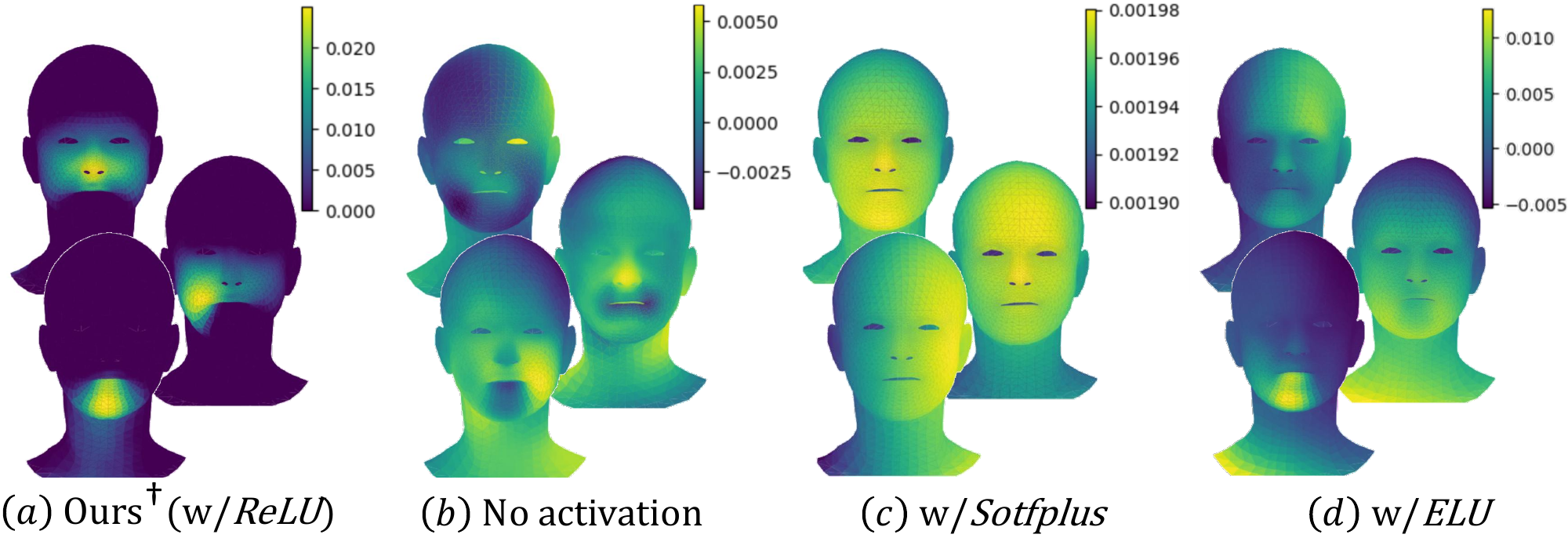}
\end{center}
\caption[Visualization of weights on COMA test data predicted from activation variants]{
    Visualization of predicted weights. Three cage vertices are randomly selected to visualize the weights predicted by each variant on the COMA test dataset.
}
\label{fig:compare_activation}
\end{figure}

Our formulation is not tied to a specific activation function. In principle, any non-negative, non-linear activation can be used to parameterize the predicted weights. To examine how the choice of activation affects non-negativity, sparsity, and deformation locality, we compared four variants for the network $\psi$: (i) \textit{ReLU} (Ours$^\dagger$), (ii) no activation, (iii) \textit{Softplus}, and (iv) \textit{ELU} with beta $0.5$. Figure~\ref{fig:compare_activation} visualizes the weights predicted by each variant. 

The \textit{Softplus} variant produced non-negative weights, with almost no zero entries. All control points therefore had nonzero contributions, although nonzero weights alone do not determine spatial locality. In this experiment, the learned supports were more diffuse than those of the ReLU variant.
This hindered the learning of localized facial deformation, leading to poor performance, as shown in Table~\ref{tab:design_study} (a).
The variants with no activation and \textit{ELU} produced sparse weights and did not ensure non-negativity. 
While these variants achieved lower inner-face errors on BIWI, they showed higher errors on COMA and Multiface. This suggests that non-negative, sparse weights are important for modeling broader, more complex facial motions.
Moreover, because of the negativity in weights, these variants produced unintended movements outside the facial region, such as in the upper chest, as shown in Figure~\ref{fig:design_study} (frame 470). This is also reflected in the MSE in $M^{\text{out}}$ as shown in Table~\ref{tab:design_study} (a).
In contrast, Ours$^\dagger$ effectively promoted non-negativity and sparsity in weights, localizing the influence of each control point, as shown in Figure~\ref{fig:compare_activation} (a). This locality led the model to reproduce the deformations with $K\text{=}512$ without generating unintended movements outside the face region. 
We therefore chose ReLU for its non-negative, potentially sparse weights and the localized support observed in these experiments; spatial locality is not guaranteed by the activation alone.

\begin{table*}[h]
\caption[Quantitative comparison of design variations]{
    Quantitative comparison of design variations. $\dagger$ indicates the base design setting with $K\text{=}512$. The best score is indicated in \textbf{Bold}.
}
\centering
\begin{tabular}{l|cccc|cccc}
    \toprule 
    \multirow{2}{*}{Design variants} & \multicolumn{4}{c|}{MSE in $M^{\text{in}}$ $\downarrow (\times 10^{-4}\,\mathrm{mm}^2)$ } & \multicolumn{4}{c}{MSE in $M^{\text{out}}$ $\downarrow (\times 10^{-4}\,\mathrm{mm}^2)$ }   \\ 
    \cline{2-9}
                           & COMA  & BIWI      & Multiface & \textit{Average} & COMA & BIWI          & Multiface  & \textit{Average}\\ 
    \hline
    Ours$^\dagger$         & \textbf{1.5242}	&	0.5278	&	\textbf{1.5461}	& \textbf{1.1994}
                                & 0.1174	&	2.0171$\times10^{-7}$	&	0.0855 & 0.0676 \\ 
    \hline
    \multicolumn{5}{l}{(a) Activation variants} \\
    \hline
    No activation          & 1.7143 &	0.5156	& 1.5636 & 1.2645
                                & 0.4973 & 1.6166$\times10^{-7}$ & 0.1015 & 0.1996 \\
    w/ \textit{Softplus}   & 10.2799 & 2.9841 & 11.5780 & 8.2807
                                & 4.2061 &  4.3432$\times10^{-7}$ & 4.7640 & 2.9900 \\
    w/ \textit{ELU}        & 1.8608  & 0.5101 &  1.6036 & 1.3248
                                & 0.3914 &  1.7556$\times10^{-7}$ & 1.0473 & 0.4796 \\
    \hline
    \multicolumn{5}{l}{(b) POU constraint} \\ 
    \hline
    w/ soft POU            & 1.8236  & 0.6003 &  1.7021 & 1.3753
                                & 0.3288 & 2.7500$\times10^{-7}$ & 0.8963 & 0.4084 \\
    \hline
    \multicolumn{5}{l}{(c) Hat function mask} \\
    \hline
    w/o mask               &  1.7916  & 0.6101 &  1.8672 & 1.4230
                                & 0.3850 &	1.6649$\times10^{-7}$ & 0.7442 & 0.3764 \\
    \hline
    \multicolumn{5}{l}{(d) Deformation representation} \\ 
    \hline
    Equation~\ref{eq:method_delta}  & 2.3562 & \textbf{0.4144} & 2.4199 & 1.7302
                                & \textbf{0.0223} & 1.6066$\times10^{-7}$ & 0.0427 & \textbf{0.0217} \\
    Equation~\ref{eq:method_matrix} & 2.4283 & 0.4449 & 2.2886	& 1.7206
                                & 0.0317 &	\textbf{1.5941$\times10^{-7}$} & \textbf{0.0391} & 0.0236 \\
    \bottomrule
\end{tabular}\label{tab:design_study}
\end{table*}

\subsubsection{Partition of unity constraint}
We approximate partition of unity through explicit row-wise normalization with a numerical-stability term, as defined in Equation~\ref{eq:pou}.
To measure the effect of this enforcement on the network $\psi$, we tested a soft constraint as an alternative. Using the pre-normalization weights $\tilde{w}$ directly for reconstruction instead of applying row-wise normalization, we applied the following loss for the soft POU constraint:
\begin{equation}\label{eq:pou_loss}
 \mathcal{L}_{\text{softPOU}}=\frac{1}{N_t}\sum^{N_t}_{j=1} \big\| \sum^K_{i=1}{\tilde{w}^{j}_{i}}-1 \big\|^2.
\end{equation}
The total loss function for the soft POU variant is $\lambda_{recon}\mathcal{L}_{recon}+\lambda_{\text{softPOU}}\mathcal{L}_{\text{softPOU}}$, where $\lambda_{recon}$ and $\lambda_{\text{softPOU}}$ are both set to $1$.
As shown in Table~\ref{tab:design_study} (b), Ours$^\dagger$, which is trained with the explicit normalization (Section~\ref{sec:weight_formulation}), produced better accuracy compared to the soft POU constraint. 
The explicitly normalized variant achieved lower inner-face error on COMA, BIWI, and Multiface than the soft POU variant.
We adopt explicit normalization based on these results. Because its denominator includes $\varepsilon$, it does not provide an exact partition-of-unity or affine-consistency guarantee.

\subsubsection{Hat function mask}
To assess the effect of the masks based on the hat function in $\mathcal{L}_{recon}$, we conducted an ablation experiment. Without the masks, the model produced deformation around the neck and upper chest region that was unrelated to the source expression, as shown in Figure~\ref{fig:design_study} (frame 470). 
Additional examples are shown in the supplementary video. Because some of the training data include global translation, the model trained without the mask tends to learn biased deformation outside the facial region, irrelevant to the source expression. The model trained with the mask mitigated this artifact and improved accuracy across most of the dataset compared with the variant without a mask, as reported in Table~\ref{tab:design_study} (c).

\subsubsection{Deformation representation}

\begin{figure*}[t]
\begin{center}
\includegraphics[width=\linewidth]{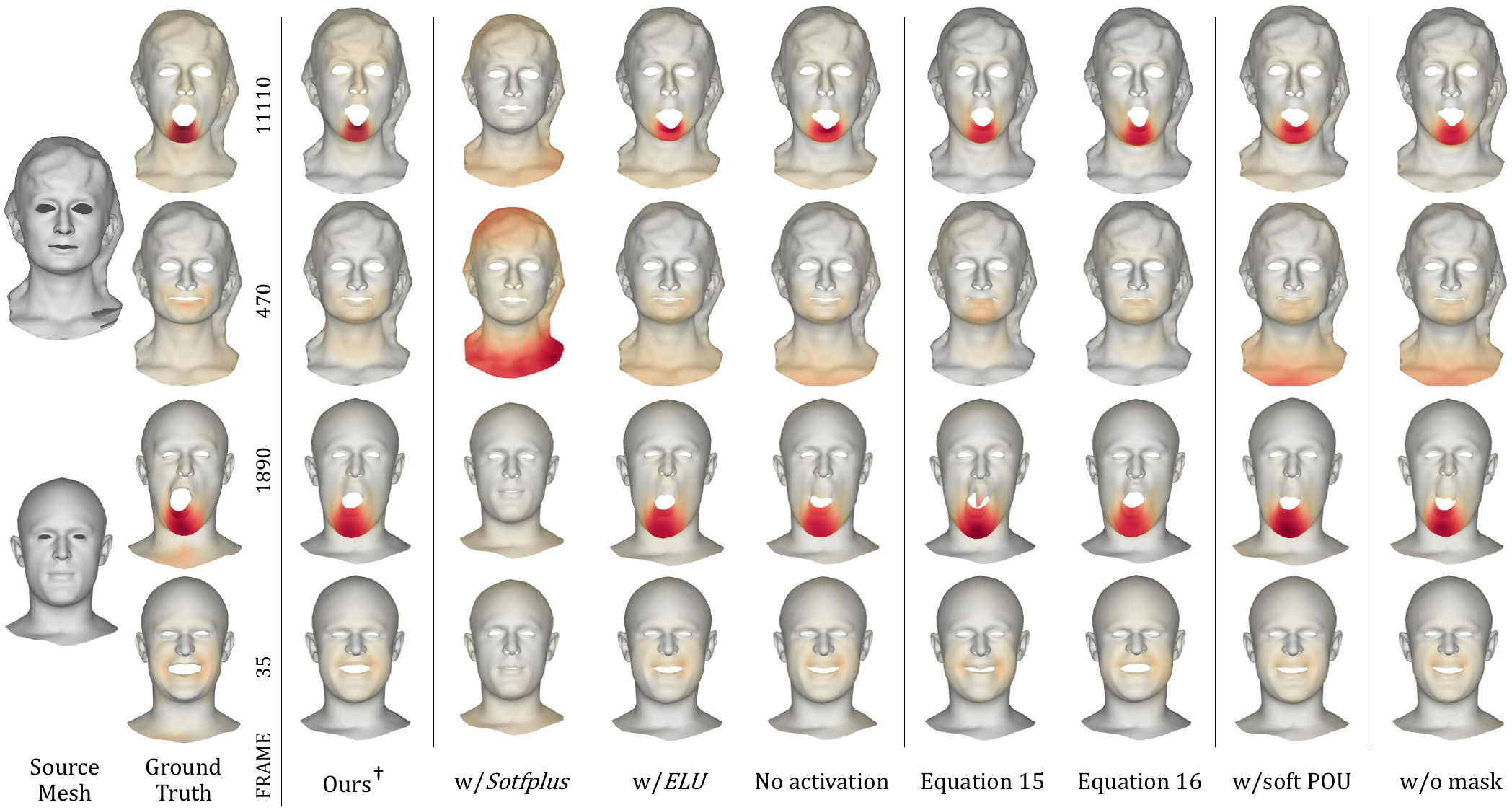}
\end{center}
\vspace{-2mm}
\caption[Visual comparison between base setting and design variants]{
    Visual comparison between the base setting and design variants using Multiface (upper) and COMA (lower). The incurred deformation on the face is colored using a yellow-orange-red color map (YlOrRd). The ground-truth data includes global translation, which produces visible deformation in the neck region.
}
\label{fig:design_study}
\end{figure*}

We compared three different representations for the network $\phi$: (i) the absolute positions of the deformed control points (Equation~\ref{eq:method_init}), (ii) the displacements of the deformed control points with respect to the initial control points, which can be expressed as follows:
\begin{equation}\label{eq:method_delta}
    \hat{p}_{\text{tgt}}^j = p_{\text{tgt}}^j + \sum_{i=1}^{K} w_i^j v_i,
\end{equation}
where $p_{\text{tgt}}^j$ is the $j$-th vertex position of the target neutral mesh and $v_i$ is interpreted as the control points displacement relative to the initial state,
and (iii) a transformation matrix that maps each initial position of the control point to its deformed position, which can be expressed as follows:
\begin{equation}\label{eq:method_matrix}
    \hat{p}_{\text{tgt}}^j = \sum_{i=1}^{K} w_i^j \ \Gamma_i \begin{bmatrix}{p^j_\text{tgt}} \\ 1 \end{bmatrix},
\end{equation}
where $\Gamma_i \in\mathbb{R}^{3\times4}$ indicates the predicted transformation matrix. 
For the third variant, the formulation is the same as linear blend skinning, except that $w$ and $\Gamma$ are defined with respect to control points rather than bones or joints. 
As shown in Figure~\ref{fig:design_study}, all three formulations produced visually similar results, demonstrating that our method is compatible with two other representations. 
The results in Table~\ref{tab:design_study} (d) indicate that using the absolute control point positions (Equation~\ref{eq:method_init}) yields the best performance in the $M^{\text{in}}$ region. In the $M^{\text{out}}$ region, Equation~\ref{eq:method_delta} achieved the best performance. Because our focus is to learn the deformation inside the $M^{\text{in}}$ region, we adopted Equation~\ref{eq:method_init} as our final formulation.

\begin{figure*}[t]
\begin{center}
\includegraphics[width=.98\linewidth]{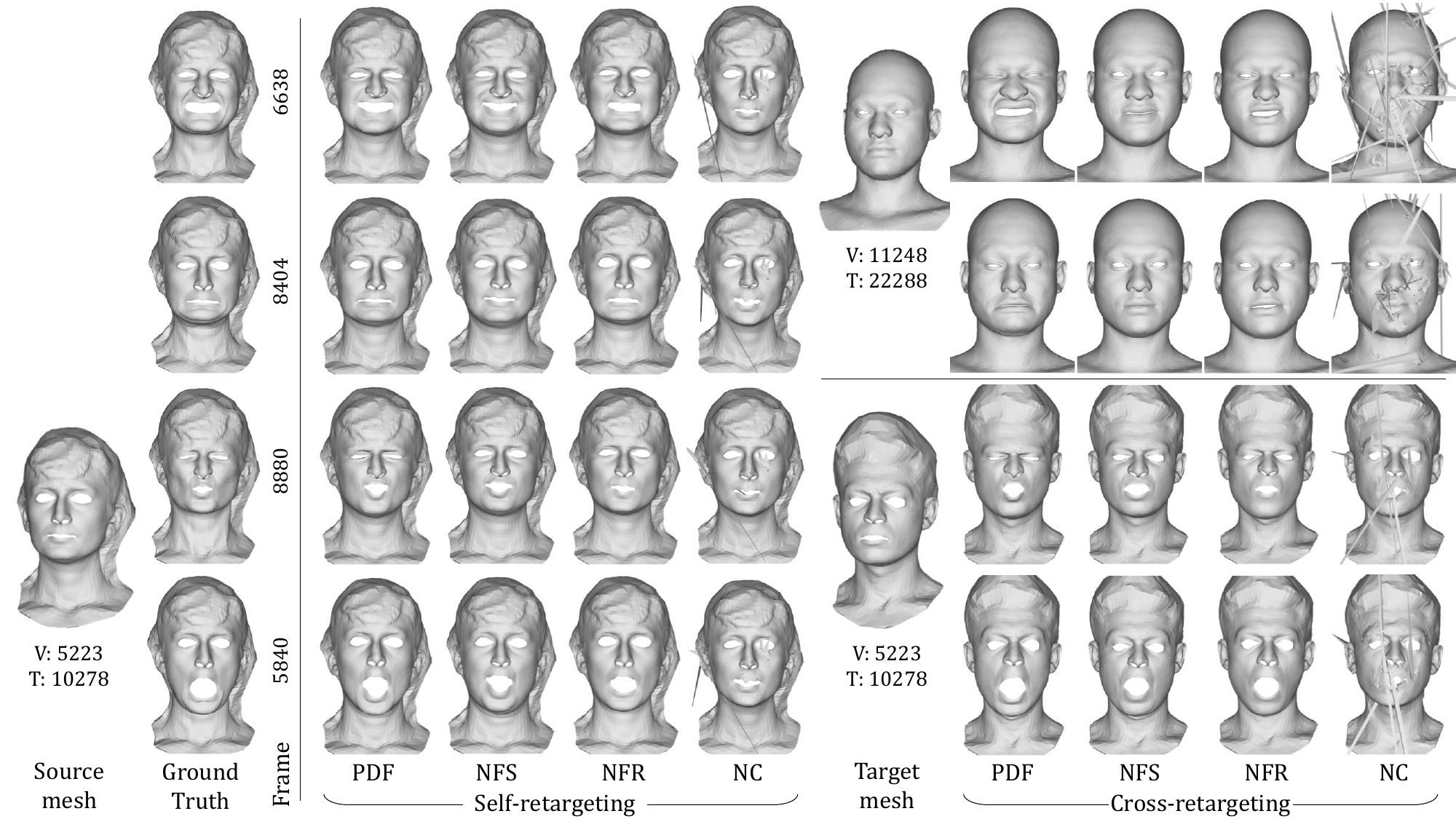}
\end{center}
\vspace{-2mm}
\caption[Visual comparison of retargeting results for Multiface test data]{
    Visual comparison of retargeting results for the Multiface test dataset produced by our method and comparative methods. The facial expressions from the source mesh are retargeted to itself, and the target meshes with different shapes and mesh structures. V and T indicate the number of vertices and triangles of the mesh, respectively.
}
\label{fig:comparison_mf}
\end{figure*}

\begin{figure}[t]
\begin{center}
\includegraphics[width=\linewidth]{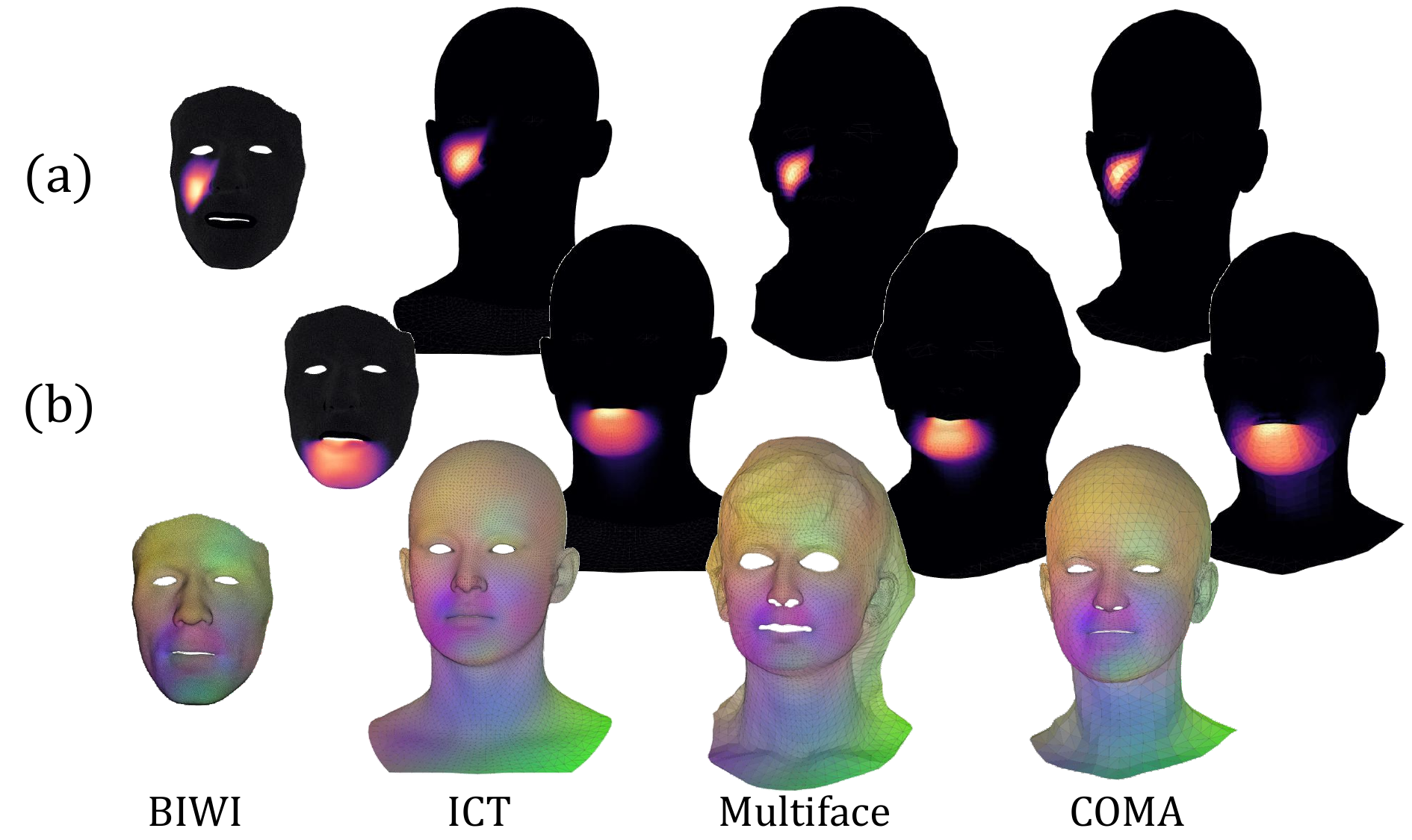}
\end{center}
\caption[Visualization of predicted weights]{
    Visualization of weights predicted from the model trained with $K\text{=}512$. The predicted weights correspond to the $413^{th}$ control point (a) and the $261^{st}$ control point (b) on test meshes from BIWI~\cite{fanelli2013biwi}, ICT~\cite{li2020ictface}, Multiface~\cite{wuu2022multiface}, and COMA~\cite{ranjan2018coma}. 
}
\label{fig:vis_coordinate}
\end{figure}

\subsubsection{Spatial Organization of Learned Weights}

The identity weight encoder $\psi$ predicts spatially coherent weights across meshes with different identities, resolutions, and geometric structures. Figure~\ref{fig:vis_coordinate} visualizes the weights predicted on test meshes from BIWI, ICT, Multiface, and COMA. For the same control point, the predicted weights are concentrated on semantically corresponding facial regions across different shapes, while maintaining local and sparse support. This consistency indicates that $\psi$ learns shape-dependent but semantically aligned weight distributions, which supports stable expression retargeting across diverse face meshes.

\begin{figure*}[t]
\begin{center}
\includegraphics[width=.98\linewidth]{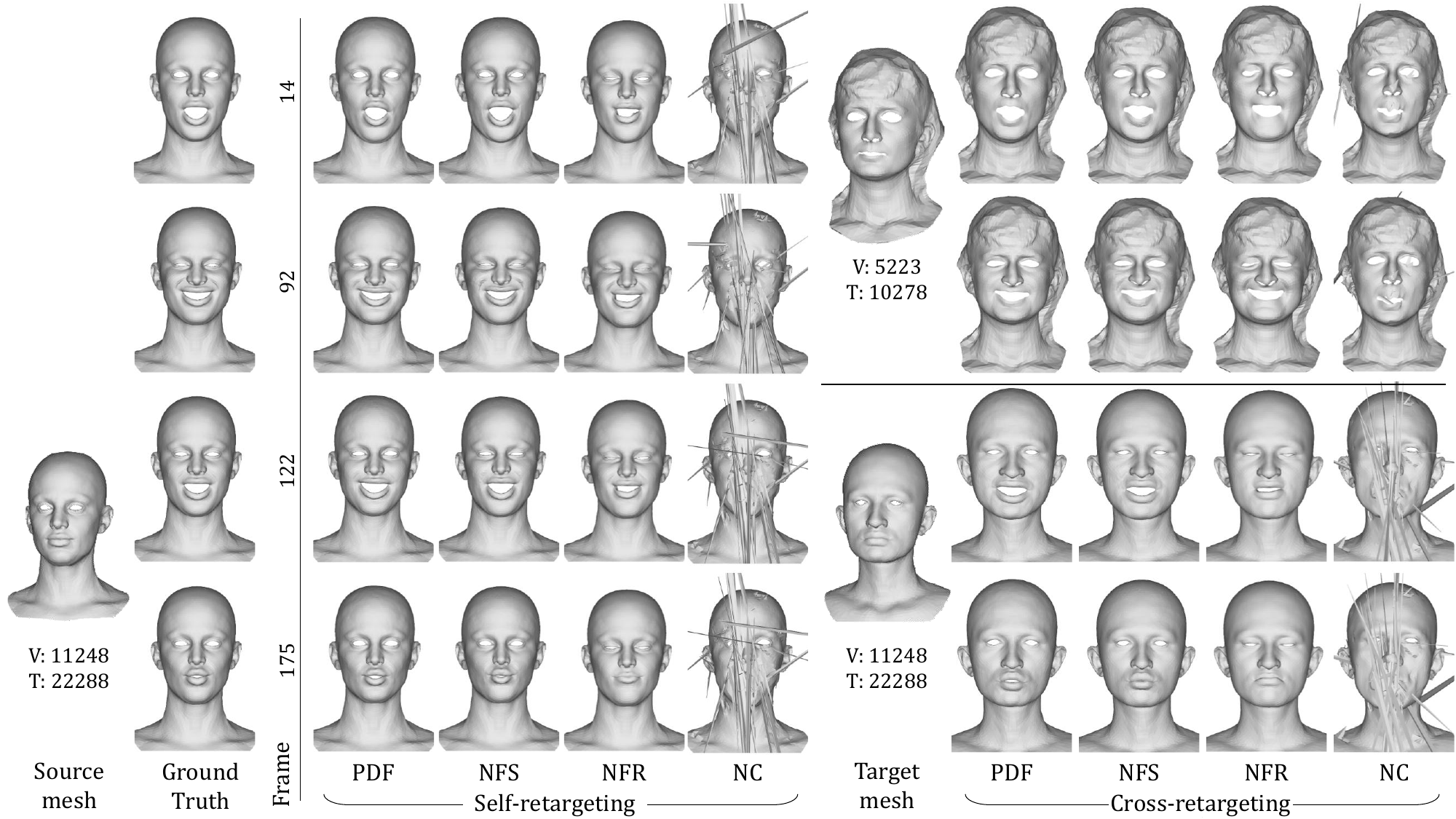}
\end{center}
\vspace{-2mm}
\caption[Visual comparison of retargeting results for ICT test data]{
    Visual comparison of retargeting results for the ICT test dataset produced by our method and comparative methods. The facial expressions from the source mesh are retargeted to itself and the target meshes with different shapes and mesh structures. V and T indicate the number of vertices and triangles of the mesh, respectively.
}
\label{fig:comparison_ict}
\end{figure*}

\begin{table*}[h]
\caption[Quantitative comparison with previous methods]{Quantitative comparison with previous methods on self-retargeting and cyclic-retargeting. Cyclic-retargeting employs ICT$\rightarrow$MF$\rightarrow$ICT and MF$\rightarrow$ICT$\rightarrow$MF cycles, in which expressions are transferred from one dataset to a test mesh in the other dataset and then mapped back to the original mesh. The best result is shown in \textbf{bold}. ICT and MF denote ICT-FaceKit and Multiface dataset, respectively.}
\centering
\begin{tabular}{lccc|ccc}
\toprule
\multirow{3}{*}{Method} 
& \multicolumn{3}{c|}{Self-retargeting} 
& \multicolumn{3}{c}{Cyclic-retargeting} \\
& \multicolumn{3}{c|}{MSE in $M^{\text{in}}$ $\downarrow (\times 10^{-4}\,\mathrm{mm}^2)$}
& \multicolumn{3}{c}{MSE in $M^{\text{in}}$ $\downarrow (\times 10^{-4}\,\mathrm{mm}^2)$} \\
\cline{2-7}
& ICT & MF & Average 
& ICT $\rightarrow$ MF $\rightarrow$ ICT & MF $\rightarrow$ ICT $\rightarrow$ MF & Average \\
\hline
NC   & 64.7652          & 11.6344          & 38.1998
     & 89.0935          & 53.2029          & 71.1482 \\
NFR  & 1.9921           & 2.9145           & 2.4533
     & 1.0957           & 4.2026           & 2.6492 \\
NFS  & 0.6161           & 2.5024           & 1.5592
     & 1.2264           & 4.1020           & 2.6642 \\
PDB & \textbf{0.5731}  & \textbf{0.8392}  & \textbf{0.7061}
     & \textbf{0.9101}  & \textbf{3.0792}  & \textbf{1.9947} \\
\bottomrule
\end{tabular}
\label{tab:compare_sota}
\end{table*}

\subsection{Comparison with Previous Methods}\label{sec:precision_qual}

We compared our method, PDB, with state-of-the-art facial animation retargeting methods, NFR~\cite{qin2023NFR} and NFS~\cite{cha2025nfs}, as well as NC~\cite{yifan2020neuralcage}, a neural cage-based deformation method. For a fair comparison, all methods were trained on ICT-FaceKit~\cite{li2020ictface} and Multiface~\cite{wuu2022multiface} using the same configuration as \citet{cha2025nfs} until convergence. For NC, we constructed a single enclosed template cage with $K$ vertices and used MVC, following the original paper. Because the Multiface dataset contains deformation outside the facial region, such as hair and upper chest motion, all errors were measured on the inner facial region using $M^{\text{in}}$.

Table~\ref{tab:compare_sota} summarizes the quantitative results. PDB achieved the best average performance in both self-retargeting and more challenging cyclic-retargeting due to repeated cross-identity transfer, demonstrating a clear margin of improvement in average error over previous methods. 
These results provide evidence that the learned factorization remains
stable under repeated cross-identity transfer on unseen test identities.

\begin{figure}[t]
\centering
\includegraphics[width=\linewidth]{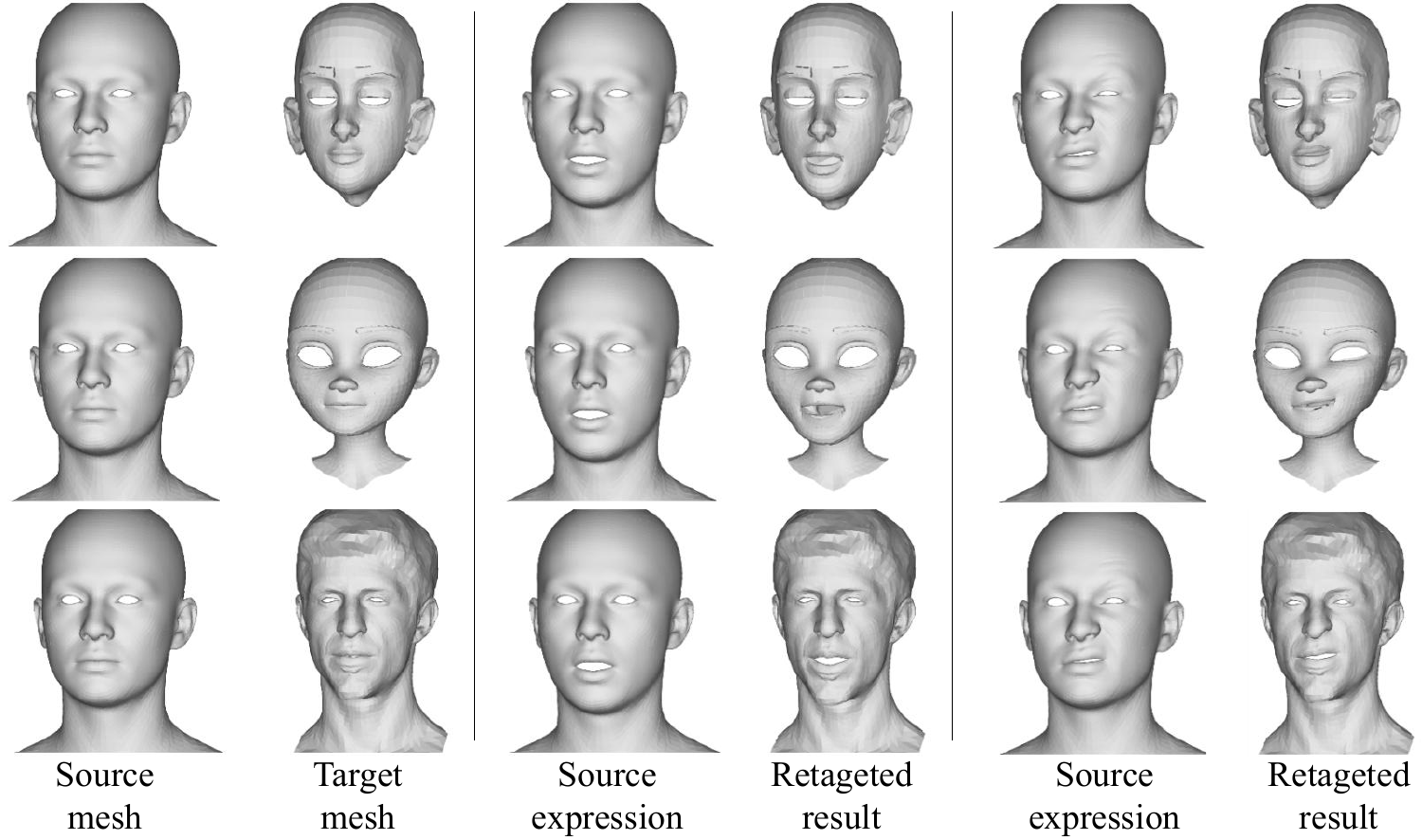}
\caption[Generalization to unseen stylized target meshes]{
    Generalization to unseen stylized target meshes. 
    PDB retargets source expressions to target meshes with substantially different facial geometry and stylized proportions.
}
\label{fig:stylized}
\end{figure}

Qualitative results are shown in Figure~\ref{fig:comparison_mf} and Figure~\ref{fig:comparison_ict}. NC failed to reproduce the target deformations. During training, NC constrains the cage to a single polyhedron and penalizes negative MVC values, which pushes the predicted cage toward a convex shape. However, many facial motions involve concave configurations, such as those around the eyes and mouth. As a result, NC did not converge properly and often produced self-intersecting cages or degenerate cage triangles, leading to severe artifacts. In contrast, PDB produced stable retargeting results without exhibiting these failure modes.
Figure~\ref{fig:stylized} further demonstrates that our method generalizes to unseen stylized target meshes with substantially different facial geometry and proportions while preserving the source expression in a stable and coherent manner.

\begin{figure}[h!]
\centering
\includegraphics[width=\linewidth]{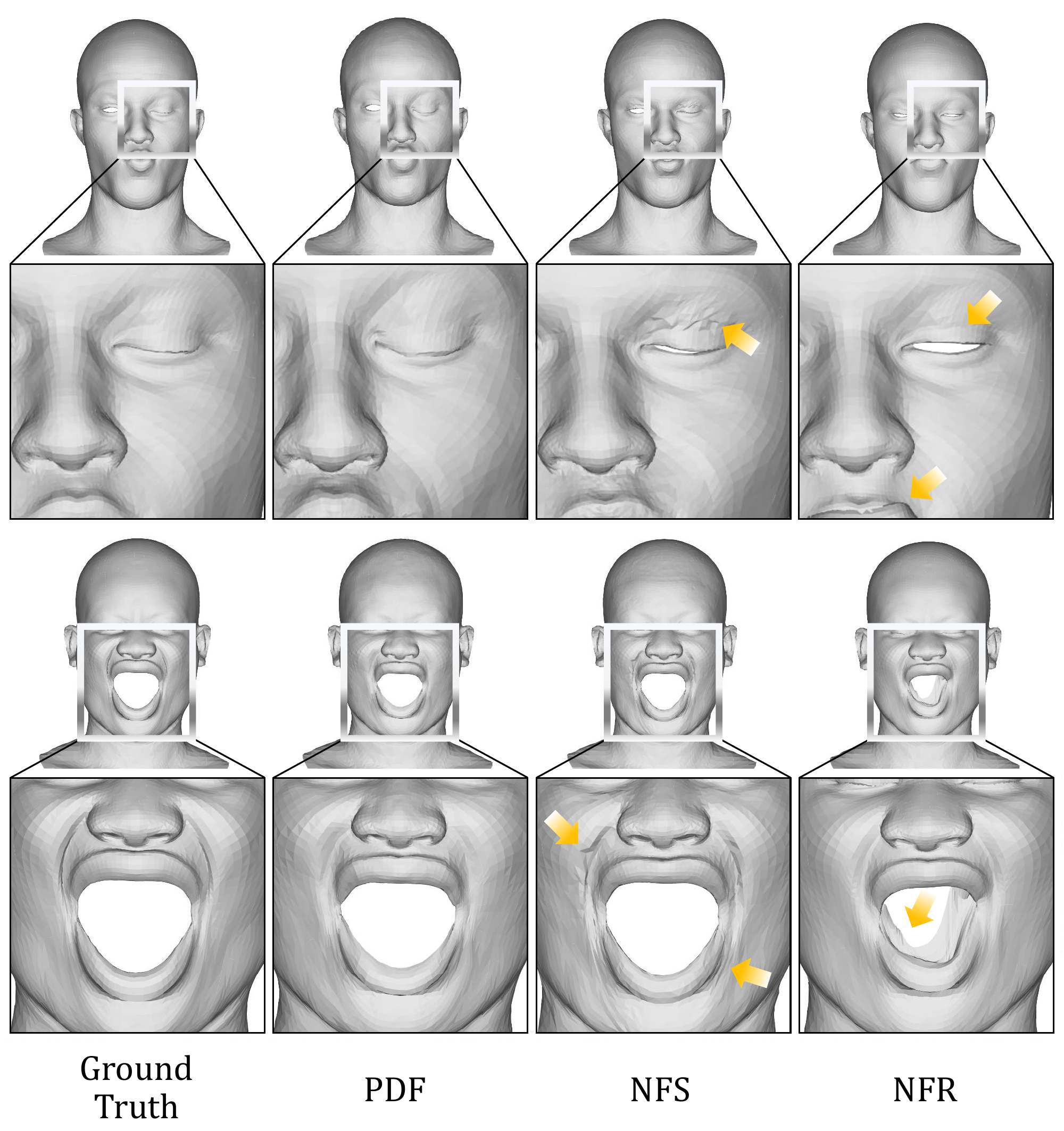}
\caption[Magnified results of the ICT test data produced by our method and comparative methods]{
    Magnified results on the ICT test dataset produced by our method and comparative methods. The meshes were rendered with flat shading to visualize the deformed surface.
}
\label{fig:comparison_ict_zoom}
\end{figure}

\subsubsection{Neutral Reconstruction}
\label{sec:identity_reproduction}
\csh{
We evaluate the special case in which the source neutral mesh, source
expression mesh, and target neutral mesh are identical. The desired
output is therefore the unchanged target neutral mesh. This experiment tests whether the learned factorization introduces drift when representing the identity deformation. It does not constitute an
analytic guarantee of linear reproduction.}
\begin{table}
\caption{Neutral reconstruction error on ICT-FaceKit and Multiface.}
\begin{tabular}{lc}
\toprule
Method & MSE in $M^{\mathrm{in}}\downarrow(\times10^{-6}\,\mathrm{mm}^2)$ \\
\midrule
NC  & $511.031$ \\
NFR & $100.984$ \\
NFS & $8.148$ \\
PDB & $8.779$ \\
\bottomrule
\end{tabular}
\end{table}



\begin{figure}[h]
\begin{center}
\includegraphics[width=\linewidth]{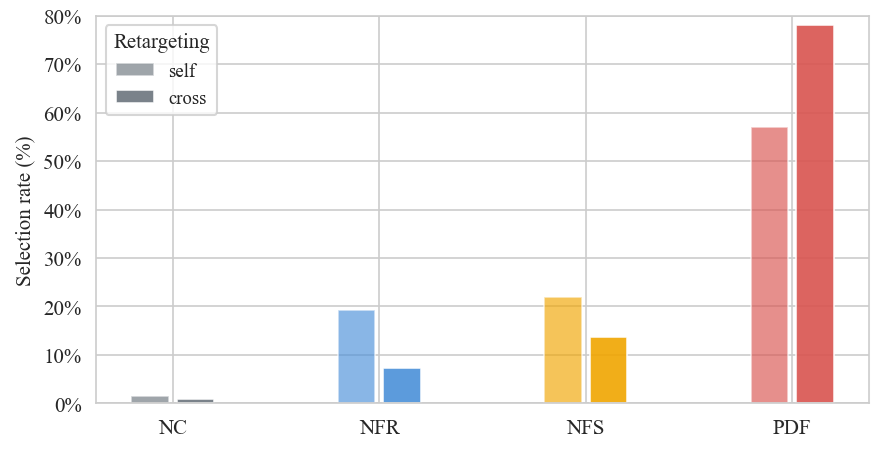}
\end{center}
\caption[User study on expression similarity]{
    User study results for expression similarity. 
    Participants selected the method that best preserved the source motion for each clip. 
}
\label{fig:external_comparison}
\end{figure}

\begin{figure}[h]
\begin{center}
\includegraphics[width=\linewidth]{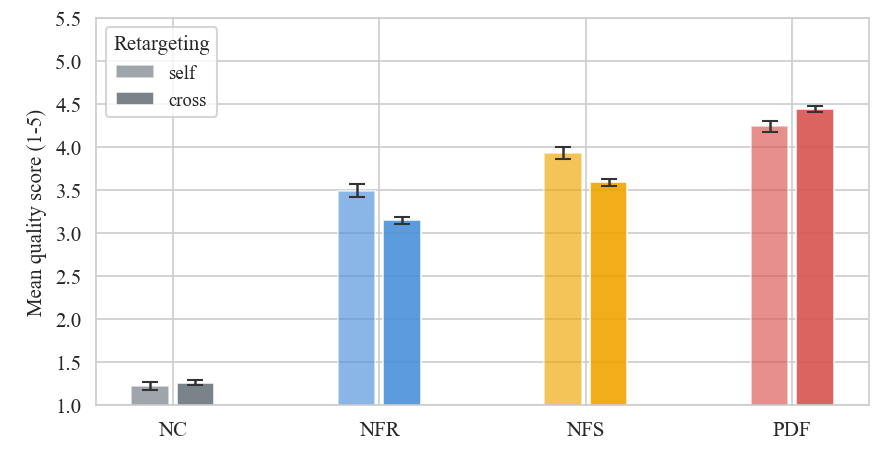}
\end{center}
\caption[User study on visual quality]{
    User study results for visual quality. 
    Participants rated the naturalness and absence of visible artifacts using a 1--5 Likert scale. 
}
\label{fig:internal_rating}
\end{figure}

\subsection{User Study for Perceptual Quality}\label{sec:user_study}

We conducted a user study with 30 participants to evaluate the perceptual quality of the retargeted animations. The participants included 12 males and 18 females, mostly in their 20s and 30s, with 14 participants having prior experience with animation or digital content creation. The study included both self-retargeting and cross-retargeting results. We used 28 video clips in total: 7 self-retargeting clips and 21 cross-retargeting clips generated from the ICT and Multiface test set.
The participants were presented with each clip and asked to evaluate two aspects: expression similarity and visual quality. Expression similarity was measured using a forced-choice selection of the method that best preserved the source motion among four randomly ordered results: PDB, NFS, NFR, and NC. Visual quality was measured using a 1--5 Likert score, where higher scores indicate more natural results with fewer visible artifacts. Figure~\ref{fig:external_comparison} and Figure~\ref{fig:internal_rating} summarize the results separately for self-retargeting and cross-retargeting.

The results show that our method was most frequently selected for expression similarity and achieved the highest visual quality scores in both self-retargeting and cross-retargeting. NC received low scores in both metrics, indicating severe visual artifacts and poor expression preservation. NFR and NFS produced visually plausible animations, but they were less effective in preserving the source expression. In contrast, ours maintained both high mesh quality and strong expression similarity, showing that the proposed representation retargets source motion faithfully while producing stable and plausible target animations.

\subsection{Local Surface Preservation}\label{sec:surface_preservation}

To evaluate local surface structure preservation, we measured the MSE between the predicted results and the ground truth in Laplacian coordinates within the inner facial region. Table~\ref{tab:compare_smooth} reports the results on the ICT and Multiface datasets. NC produced very large errors on the ICT data, indicating unstable local surface behavior. Among the remaining methods, NFR yielded the lowest errors overall while PDB achieved the second-best result, outperforming NFS on both datasets with a substantially lower average error. 
NFR achieved strong surface regularity through a global linear solve, but local prediction noise could propagate globally and cause retargeting errors.
This tendency is also observed in the qualitative comparison presented in Figure~\ref{fig:comparison_ict_zoom}. In the magnified eyelid and mouth regions, our method produced a smooth surface that closely matches the ground truth, whereas NFS exhibited noisy local surface artifacts. NFR preserved the smooth surface but produced undesired deformation in both the eyelid and the inner mouth, indicating global entanglement.
These results indicate that the proposed point-based factorization preserves local surface structure more effectively than per-vertex prediction, while avoiding the severe instability observed in NC.


\begin{table}[t]
\caption[Quantitative comparison of local surface structure preservation]{Quantitative comparison of local surface structure preservation, measured by the MSE between \emph{Laplacian coordinates} within the inner facial region. The best result is shown in \textbf{bold}, and the second-best result is \underline{underlined}.}
\centering
\begin{tabular}{lccc}
\toprule
\multirow{2}{*}{Method} & \multicolumn{3}{c}{MSE in $M^{\text{in}}$  $\downarrow$ $(\times 10^{-4}\,\mathrm{mm}^2)$} \\
\cline{2-4}
& ICT & Multiface & Average \\
\hline
NC   & 5927.2172            & 178.2116            & 3052.7144            \\
NFR  & \textbf{1.1284}      & \textbf{1.2206}     & \textbf{1.1745}      \\
NFS  & 1.6646               & 3.9028              & 2.7837               \\
PDB & \underline{1.2771}   & \underline{1.8993}  & \underline{1.5882}   \\
\bottomrule
\end{tabular}
\label{tab:compare_smooth}
\end{table}

\subsection{Runtime Comparison}

We compared the runtime of our method with state-of-the-art retargeting methods on test performance sequences.
For this purpose, a single performance (EXP\_free\_face) from the Multiface dataset and custom human performances captured with \textbf{Live Link Face}~\cite{livelink} and ICT-FaceKit~\cite{li2020ictface} (capture \#1, capture \#2), were employed. All methods were evaluated on the same hardware, a single NVIDIA RTX A5000 24GB GPU. The reported time includes the end-to-end runtime for each sequence, including any one-time preprocessing required for each target mesh.
NFR and NFS both require mesh-dependent preprocessing, such as computing the Laplace--Beltrami operator.
This required additional computation time, thereby increasing latency by around 15 seconds.
In contrast, PDB does not require any such preprocessed information and directly predicts the retargeted mesh.
Moreover, unlike NFR and NFS, which rely on a neural decoder to map latent to deformation, PDB produces the deformation through a simple matrix product between predicted weights and deformed control points, reducing both computation and runtime.
Table~\ref{tab:compare_mf_ROM_time} shows that PDB has the lowest end-to-end runtime for all three sequences. These measurements include target-specific setup and do not separately quantify preprocessing and per-frame inference.

\begin{table}[t]
\caption[Runtime comparison]{
    Runtime comparison for retargeting complete performance sequences. Total time includes one-time target-specific preprocessing and processing all frames.
}
\centering
\begin{tabular}{lcccc}
    \toprule 
    \multirow{2}{*}{Data} & \multirow{2}{*}{Frames} & \multicolumn{3}{c}{Time (min:sec)} \\
    \cline{3-5}
                           &        & NFR & NFS & PDB \\ 
    \hline 
    
    ICT (capture \#1)        & 183    & 00:25 & 00:17 & \textbf{00:01}  \\
    ICT (capture \#2)        & 1,147  & 01:19 & 00:34 & \textbf{00:06}  \\
    Multiface (test data)    & 11,649 & 17:32 & 02:29 & \textbf{00:50}  \\

    \bottomrule
\end{tabular}\label{tab:compare_mf_ROM_time}
\end{table}

\section{Limitation}

High-frequency components of the deformation, such as wrinkles or abrupt movements, may not be fully captured by our method. 
The model tends to place control points in regions with large deformations, such as the eyes and mouth. Because wrinkles are subtle and vary across identities, these fine details are often overlooked.
Increasing the number of control points or introducing additional constraints on their spatial support may improve fine-detail reconstruction, but this remains to be evaluated.
Increasing $K$ reduces the compactness of the representation and increases the storage and multiplication cost of the $N_t\times K$ weight matrix. Even when $K=N_t$, the learned weighted reconstruction is not generally equivalent to independent per-vertex displacement prediction. In this work, we used $K=512$ as a heuristic setting rather than an optimized value. As shown in Table~\ref{tab:number_of_k}, the model also trained successfully with $K=256$, achieving comparable performance to $K=512$ on self-retargeting. 
While this suggests that the method is not tied to a specific value of $K$, identifying the optimal number of control points for faithfully reproducing fine details remains an important direction for future work.

\begin{table}[t]
\caption[Quantitative comparison with different numbers of control points]{
    Quantitative comparison with different numbers of control points $K$ on self-retargeting. The best result is shown in \textbf{bold}.
}
\centering
\begin{tabular}{lccc}
\toprule
\multirow{3}{*}{Control points $K$}
& \multicolumn{3}{c}{Self-retargeting}  \\
& \multicolumn{3}{c}{MSE in \(M^{\text{in}}\) \(\downarrow\) \((\times 10^{-4}\,\mathrm{mm}^2)\)}\\
\cline{2-4}
& ICT & MF & Average  \\
\hline
$K$=512 & \textbf{0.5731} & 0.8392 & 0.7061 \\
$K$=256 & 0.5820 & \textbf{0.8025} & \textbf{0.6922} \\
\bottomrule
\end{tabular}
\label{tab:number_of_k}
\end{table}
\section{Conclusion}

We presented PDB, a method for mesh-agnostic facial animation retargeting that combines source-dependent deformed control points with target-dependent blending weights. The target mesh is reconstructed by a matrix product, without a predefined cage, precomputed coordinates, a learned per-element deformation decoder, or a global reconstruction solve. The learned non-negative weights exhibit localized support in our evaluations. Self-retargeting and cyclic-retargeting experiments show lower reconstruction and round-trip errors than the compared methods, while qualitative cross-identity results and perceptual evaluations support expression fidelity and visual quality. Additional analysis shows improved local surface preservation relative to the evaluated displacement method, transfer to unseen stylized meshes, and shorter end-to-end sequence processing times. Capturing fine-scale details and selecting the number of control points remain directions for future work.





\bibliographystyle{ACM-Reference-Format}
\bibliography{sample-base}

\clearpage
\section*{Supplementary Material}

\begin{figure}[t]
\begin{center}
\includegraphics[width=\linewidth]{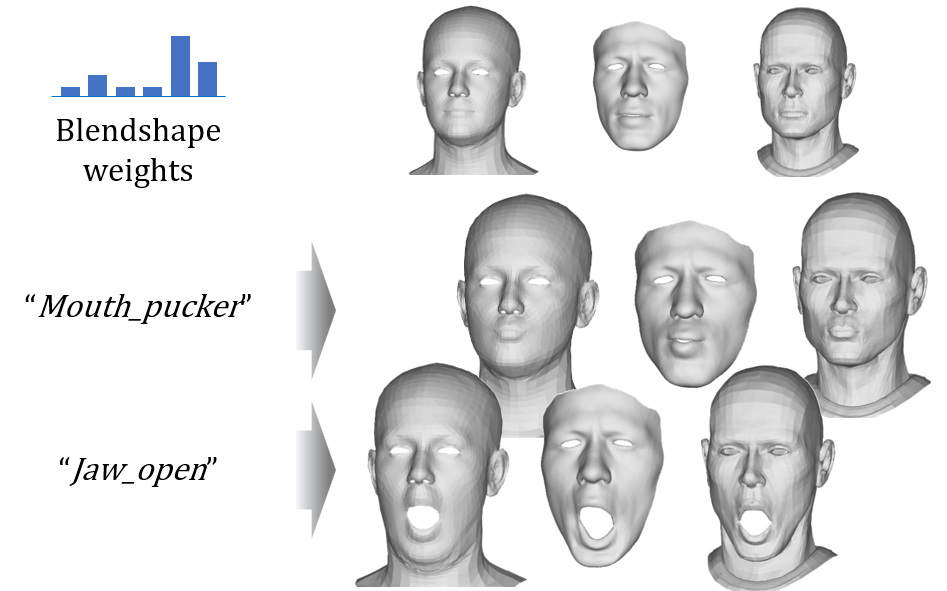}
\end{center}
\caption[Expression controlled by blendshape coefficients]{
    Expression controlled by blendshape coefficients using the model with latent aligned to the ICT blendshape. (Face meshes from left to right: FLAME~\cite{li2017flame}, BIWI~\cite{fanelli2013biwi}, and DT~\cite{sumner2004deformation}.)
}
\label{fig:blendshape_control}
\end{figure}

The control space for the control points can be provided by enforcing the latent of $\phi$ to align with the blendshape weights, following~\citet{qin2023NFR}.
Figure~\ref{fig:blendshape_control} shows an example of our model trained with blendshape alignment, providing facial expression editing via latent manipulation.
To enable the intuitive control via latent manipulation, we align the latent of $\phi$ with the target blendshape coefficients using an additional loss as follows:
\begin{equation}\label{eq:loss_blendshape}
    \mathcal{L}_{\text{BS}} \;=\; \big\| z_{\phi} - z_{\text{BS}} \big\|_2^2,
\end{equation}
where $z_{\phi}$ is the latent of $\phi$ and $z_{\text{BS}}$ is the target blendshape coefficient vector. The total loss is $\lambda_{\text{total}}\mathcal{L}_{\text{total}} + \lambda_{\text{BS}}\mathcal{L}_{\text{BS}}$, where $\lambda_{\text{total}}$ and $ \lambda_{\text{BS}}$ are set to $1$ and $0.5$, respectively.

\section{Training Data Preparation}\label{sec:data_prepare}

During training, the network input is a per-vertex feature formed by concatenating the vertex position and vertex normal (row-wise). 
For the design study, public benchmark datasets—COMA~\cite{ranjan2018coma}, BIWI~\cite{fanelli2013biwi}, and Multiface~\cite{bickel2007multi} were used for training the model. Each dataset was created by capturing human facial expressions, following a similar capture protocol: sequences start from a neutral face, perform an expression, and return to neutral. As a result, most frames are mostly neutral. To reduce duplicate neutral expressions, Principal Component Analysis (PCA) was applied to each captured expression sequence, and the top principal components that collectively explain at least 95\% of the total variance were retained.
Then, data was sampled uniformly from each expression PCA basis for training.

The same mesh can be represented with a varying number of vertices. To make the model output a consistent result regardless of discretization, we augmented the data by subsampling mesh vertices: for each mesh, we randomly removed a number of vertices ranging from 200 up to one-sixth of the total vertices of the mesh.

\begin{figure}[h]
\begin{center}
\includegraphics[width=\linewidth]{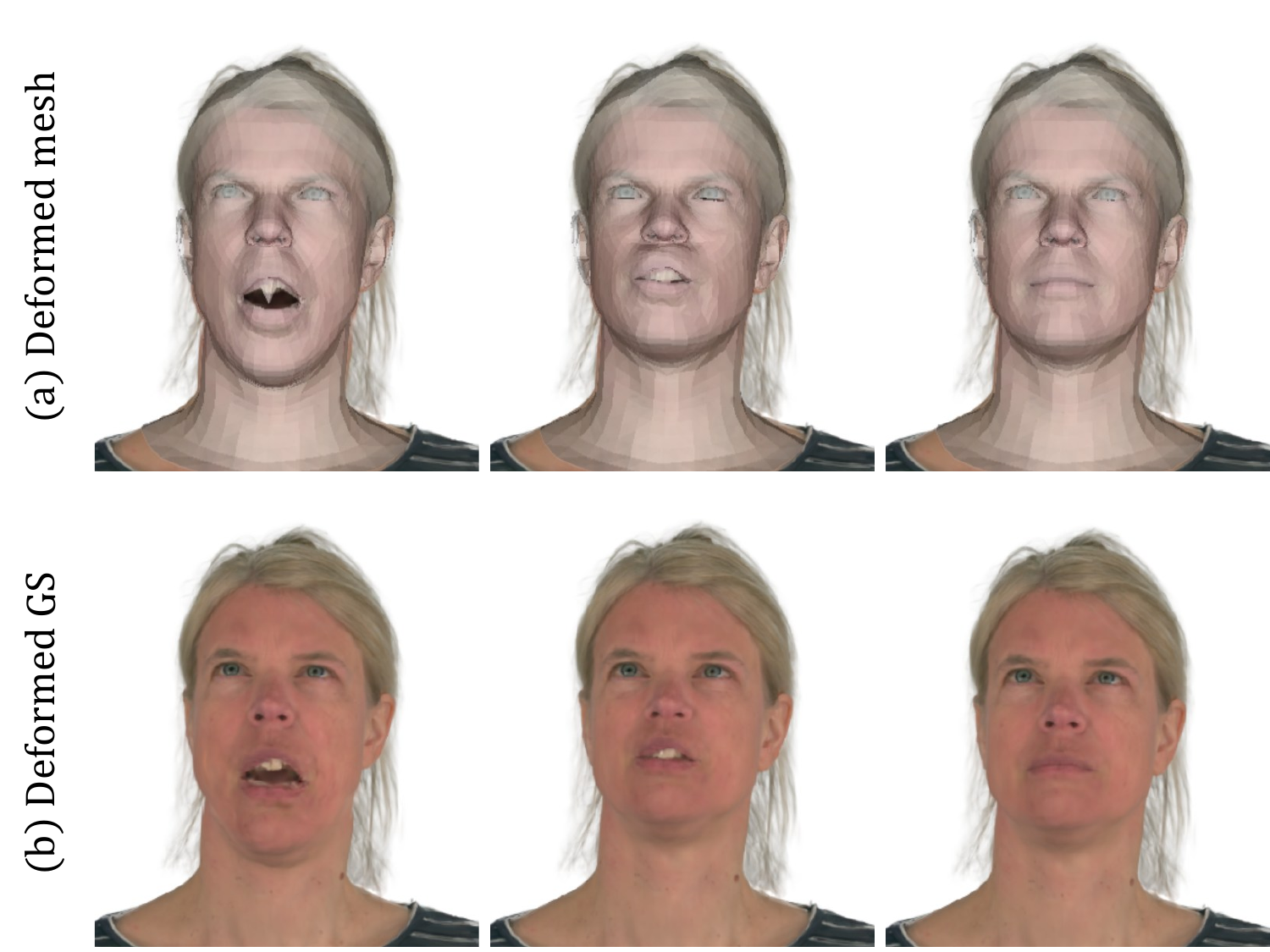}
\end{center}
\caption[Application on GaussianAvatar]{
    Application on GaussianAvatar~\cite{qian2024gaussianavatars}. The deformation applied to the underlying mesh (a) and the corresponding rendered results (b). 
}
\label{fig:application}
\end{figure}

\section{Application}

Our method can be applied not only to meshes but also to point clouds and Gaussian splats (GS) enclosed by the learned control points. For application to these targets, we first align the target’s orientation and scale, apply the deformation, and then restore the original orientation. For alignment, we coarsely match the target scale to that of a training mesh, then define approximate correspondences to perform Procrustes analysis. Figure~\ref{fig:application} shows an example of our method applied to the GS-based method.

\begin{figure}[h]
\centering
\includegraphics[width=\linewidth]{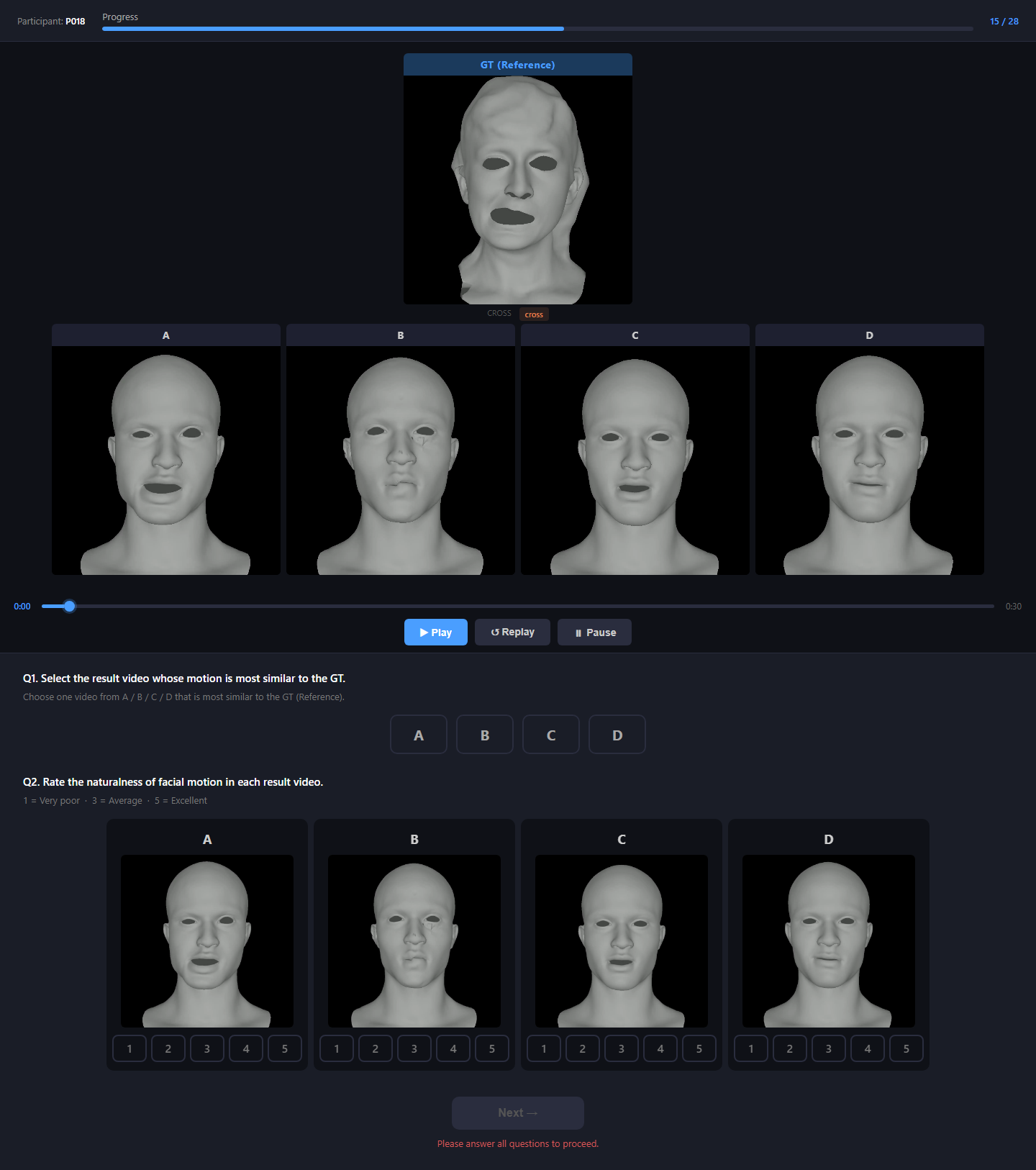}

\caption[A sample trial of the user study]{
    A sample trial of the user study.
}
\label{fig:sample_query}
\end{figure}

\end{document}